\documentclass[acmsmall]{acmart}
\usepackage{natbib}
\usepackage{{graphicx}}
\usepackage{subfigure}
\usepackage{float}
\usepackage{multirow}
\usepackage{array}
\usepackage{setspace}

\theoremstyle{definition}
\newtheorem{definition}{Definition} 
\begin{document}

\title{Privacy and Fairness in Federated Learning: on the Perspective of Trade-off}

\author{Huiqiang Chen, Tianqing Zhu*, Tao Zhang}
\email{{huiqiang.chen, tao.zhang-3}@student.uts.edu.au, tianqing.zhu@uts.edu.au}
\affiliation{%
  \institution{University of Technology Sydney}
  \streetaddress{PO Box 123 Broadway}
  \city{Sydney}
  \state{NSW}
  \country{Australia}
  \postcode{2007}
}

\author{Wanlei Zhou}
\email{wlzhou@cityu.edu.mo}
\affiliation{%
  \institution{City University of Macau}
  \streetaddress{Avenida Padre Tomás Pereira Taipa}
  \city{Macau}
  \country{China}
}
\author{Philip S. Yu}
\email{psyu@uic.edu}
\affiliation{%
  \institution{University of Illinois at Chicago}
  \streetaddress{851 S. Morgan St., Rm 1138 SEO, Chicago, IL 60607}
  \city{Chicago}
  \country{US}
}

\begin{abstract}
Federated learning (FL) has been a hot topic in recent years. Ever since it was introduced, researchers have endeavored to devise FL systems that protect privacy or ensure fair results, with most research focusing on one or the other. As two crucial ethical notions, the interactions between privacy and fairness are comparatively less studied. However, since privacy and fairness compete, considering each in isolation will inevitably come at the cost of the other. To provide a broad view of these two critical topics, we presented a detailed literature review of privacy and fairness issues, highlighting unique challenges posed by FL and solutions in federated settings. We further systematically surveyed different interactions between privacy and fairness, trying to reveal how privacy and fairness could affect each other and point out new research directions in fair and private FL.

\end{abstract}


\maketitle

\section{Introduction}
\label{sec: intro}
Machine learning has changed our lives and will undoubtedly bring us more excitement. However, its success is closely tied to the availability of large-scale training data, and as new learning models keep emerging, the demand for more data persists relentlessly. One worrisome issue with collecting massive amounts of data is the risks that present to privacy. FL \cite{mcmahan2017communication} has emerged as an attractive learning paradigm to meet privacy requirements. 

Unlike traditional centralized machine learning, FL trains models in a distributed and parallel way such that different clients collectively train a model with their training data. This technique offers two enormous benefits. First, it saves companies the costly process of collecting large-scale data because the clients provide their local data. Second, it preserves the client's privacy by keeping data locally. With such benefits, it is no surprise that the industry has already leaped to put FL into practice, such as Gboard \cite{hard2018federated}.

\subsection{Privacy and Fairness in FL}
Great achievements have been made with FL. However, the paradigm of FL still suffers from ethical issues surrounding data use. One of those ethical issues is privacy. Although raw data never leave the device, the uploaded gradients/parameters still carry local information. Therefore, a trained model could hold the client’s data distribution. Consequently, an adversary can infer information about what data are included in the training set \cite{hitaj2017deep,9148790} or, even worse, reconstruct the training data \cite{geiping2020inverting}. As such, the research community is endeavoring to identify all potential privacy risks by launching different privacy attacks \cite{8835245}. Accordingly, defenses for all these attacks are also being proposed to secure private data \cite{wei2020federated}. This wargaming between attack and defense is leading us to more private FL environments. 

Another ethical issue in FL is fairness, which refers to reducing the model’s bias towards disadvantaged groups, such as ethnic minorities, women, or the aged. Fairness in FL is defined at two different levels. The first pertains to \textit{algorithmic fairness} \cite{mehrabi2021survey,chouldechova2018frontiers}, where model output should not skew towards disadvantaged groups defined by some sensitive attributes. The second pertains to \textit{client fairness} \cite{li2019fair,mohri2019agnostic}. In vanilla FL \cite{mcmahan2017communication}, models trained on the larger dataset are given higher importance during aggregation. Hence, the global model will be optimized to capture the data distributions of clients with a larger dataset. Therefore the model performance will vary significantly among clients, which imposes unfairness at the client level.

Privacy and fairness are two crucial ethical notions, and violating either of them is unacceptable. However, to date, the research community has primarily considered these two issues separately, yet they are inextricably entwined. For example, it is well known that privacy comes at the cost of accuracy. What is surprising is that the cost is not consistent across all groups as expected, where disadvantaged groups often suffer more of an accuracy decrease than the other groups due to data scarcity \cite{bagdasaryan2019differential, Kuppam2020FairDM}. In other words, ensuring privacy can exacerbate the inequities between groups. Fairness, in turn, may negatively affect privacy. For example, to ensure a classification model is fair, the server usually needs to know the underlying distribution of the training dataset to eliminate bias existing in either the training data \cite{feldman2015certifying,kamiran2012data,kamiran2010classification} or the model \cite{zafar2017fairness,goh2016satisfying,zemel2013learning}. This means the client will share more data with the server, increasing the privacy risk.

Therefore, in addition to reviewing privacy and fairness issues in FL, another motivation of this survey is to explore the possible interactions between privacy and fairness and to discuss the relationships between fairness and privacy. It is worth noting that the issue of client fairness adds an extra layer of complexity to the federated setting. To the best of our knowledge, this survey is the first attempt to examine the relationships between privacy and fairness in the federated setting.
\begin{table}[htp]
\centering
\caption{Comparison to Related Surveys on Privacy or Fairness in FL}
\label{tab:Survey Comparison}
\small
\begin{tabular}{cccccc}
\toprule
\multirow{2}{*}{Reference} & \multicolumn{2}{c}{Privacy-preserving} & \multicolumn{2}{c}{Fairness-aware} & \multirow{2}{*}{\begin{tabular}[c]{@{}c@{}}Interactions \\ between privacy \\ and fairness\end{tabular}} \\ \cline{2-5}
 & \begin{tabular}[c]{@{}c@{}}Privacy \\ attack\end{tabular} & Defense & \begin{tabular}[c]{@{}c@{}}Algorithmic \\ fairness\end{tabular} & \begin{tabular}[c]{@{}c@{}}Client \\ fairness\end{tabular} &  \\ 
\midrule
\cite{yang2019federated} & \checkmark & \checkmark &  &  &  \\
\cite{yin2021comprehensive} & \checkmark & \checkmark &  &  &  \\
\cite{lyu2020threats} & \checkmark & \checkmark &  &  &  \\
\cite{kairouz2019advances} & \checkmark & \checkmark & \checkmark & \checkmark &  \\
Our work & \checkmark & \checkmark & \checkmark & \checkmark & \checkmark \\ 
\bottomrule
\end{tabular}
\end{table}

\subsection{Main Contribution}
This survey provides a broad view of privacy and fairness issues in FL. We first illustrated that FL is not as private as it claimed to be. Adversarial clients and server have several new attack vectors at their disposal, and we outlined several techniques for preserving privacy against these attacks. Turning to fairness, we explained the two lines of fairness notions adopted in FL and the corresponding debiasing strategies. Lastly, we discussed interactions between privacy and fairness. Our contributions are as follows:
\begin{itemize}
\item This is the first survey that provides a comprehensive overview of privacy, fairness, and the interactions between the two. 

\item We present a detailed survey of privacy attacks and defenses in FL, discuss how these privacy attacks could damage privacy in FL and highlight the assumptions and principle methods of these attack strategies. 

\item Following a rigorous enumeration of the sources of bias in the FL pipeline, we discuss the fairness notions adopted in FL and summarize fairness-aware FL approaches.

\item We point out several future research directions toward training private and fair FL models.
\end{itemize}

\section{Background Knowledge}
\label{sec: background}
\subsection{Definition of FL}
The goal of FL is to train a global model in a distributed way. The objective function is formulated as: 
\begin{equation}
    \mathop {\min }\limits_w f\left( w \right) = \sum\limits_{k = 1}^m {{p_k}{F_k}\left( w \right)} \label{FL object func}
\end{equation}
where $m$ is the number of clients, $p_k > 0$ is the aggregating weight of client $k$, satisfying $\sum_k p_k = 1$. $F_k(w)$ is the empirical risk on client $k$'s data. In a trivial setting, $p_k$ is the ratio of local samples to total samples of all clients. The process of FL consists of two stages: the training and inference stages. Three actors are involved: 1) clients, each of which has a local dataset and will use it to contribute to the global model’s training by uploading local gradients/parameters, noting that each client’s dataset may vary from the others; 2) a server that coordinates the learning process; and 3) users, who will use the final well-trained model. 

In each iteration of FL, selected clients download the global model and perform learning algorithms locally. They communicate their updates to the server for aggregation and model updating. This interaction between clients and the server repeats until the model converges. At the inference stage, a well-trained model is deployed to users, where users can infer the model via black-box access. This step is no different from a traditional data center approach.


\subsection{Privacy Disclosure in FL}
\begin{figure}[htp]\centering
    \subfigure[Overview of reconstruction attack in FL. An adversary iterative optimizes generated sample by matching gradients $\nabla W$.]{
    \includegraphics[scale=0.4]{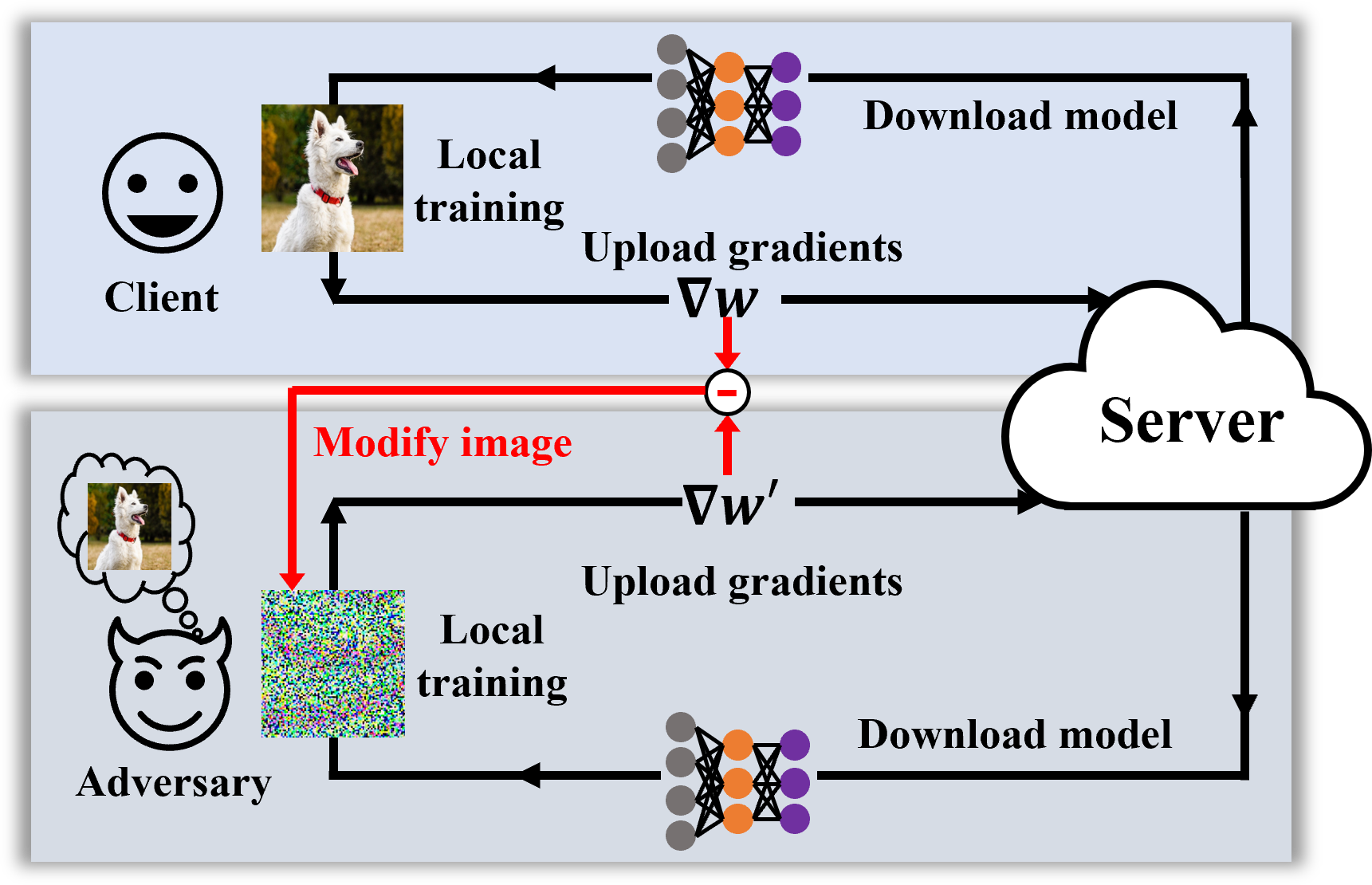}        \label{fig:DLG}}
    \hspace{4mm} 
    \subfigure[Attack results on MNIST, CIFAR-100, SVHN and LFW. The adversary is able to reconstruct the private images of other clients truthfully (reproduced from \cite{zhu2019deep}).] {
    \includegraphics[scale=0.4]{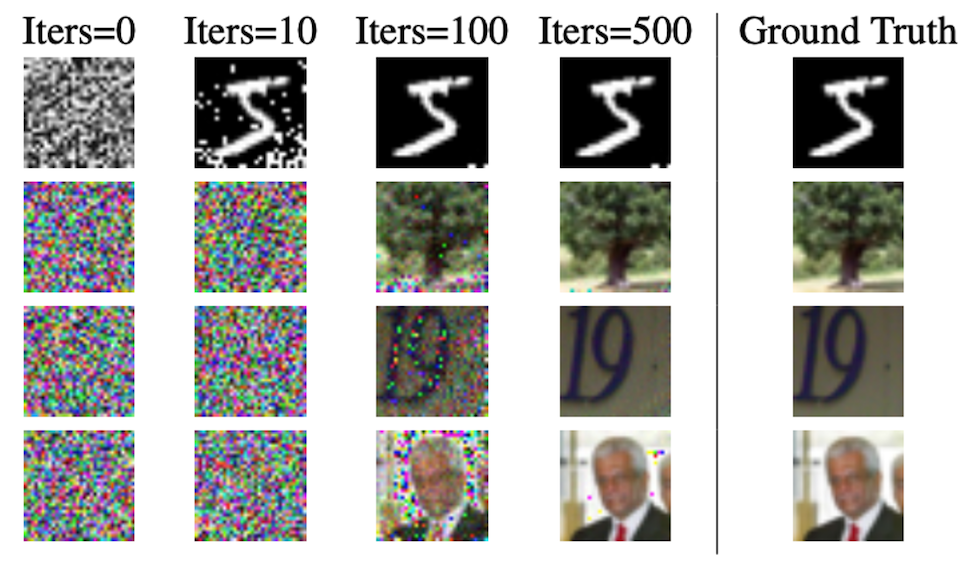}        \label{fig:ReconAttack result}}
    \caption{Reconstruction attack in FL.}
    \label{fig:Privacy Disclosure}
\end{figure}

Recent research verified the privacy risk of FL. Adversaries can glean the participants’ training data. For example, Zhu et al. \cite{zhu2019deep} fully reconstructed training data from the victim client through the uploaded gradients as shown in Fig.\ref{fig:DLG}. In the course of the FL pipeline, several attack interfaces exist for an adversary, who could be the server or a client in FL. The attack could occur during the training or inference stage, within or outside the FL. The attack targets include membership, property, class representative, and the raw data \cite{yin2021comprehensive}. 

\subsubsection{Membership Inference Attacks}
Membership Inference Attacks (MIA) aim to identify whether a given sample was used to train the target model. These types of attacks can pose privacy risks to individuals. For example, confirming a patient’s clinical record was used to train a model associated with a particular disease would reveal that patient’s health condition. MIA was initially investigated by Shokri et al. \cite{shokri2017membership}. The attack models are essentially binary classifiers. Given an instance $\boldsymbol{X}$ and a target model $F_t$, the goal of the MIA model is to identify whether or not  $\boldsymbol{X}$ is contained within the training dataset $D$ of the target model $F_t$.

\subsubsection{Property Inference Attacks} 
 Property inference attacks aim to recover some property of the training set, which may be irrelevant to the main tasks. Such as the property of "wearing glasses" against a gender classifier or the composition of the training dataset. This kind of attack also leads to privacy issues. With proper prior knowledge, the adversary can infer the presence of a specific sample in the training set.

\subsubsection{Model Inversion Attacks}
Model inversion attacks aim to recover class-specific features or construct class representatives by accessing the target model and other possible auxiliary information. The recovered data is a representing sample (usually a synthetic sample) that only reflects some aspects of the training data and is not a member of the training set.

\subsubsection{Reconstruction Attacks}
Reconstruction attacks \cite{dinur2003revealing} aim to reconstruct a probabilistic version of samples in the training set. Success in a reconstruction attack is measured by comparing the reconstruction with the original data. If the two are similar, the attack has been successful. Fig.\ref{fig:ReconAttack result} \cite{zhu2019deep} shows an example. Unlike the model inversion attacks, the recovered data here is almost the same as the training dataset at the pixel level and belongs to the training dataset.

\subsection{Privacy-preserving Techniques}
In the realm of FL, plenty of studies have shown how to break the basic privacy assurances, such as determining a client’s membership in the training set  \cite{8835245}, ascertaining the class representations of the client’s training data \cite{hitaj2017deep,wang2019beyond, song2020analyzing}, and, the worst case of all, procuring the raw training data \cite{geiping2020inverting}. Several privacy-preserving techniques can help stop these privacy leakages, including cryptographic techniques and the perturbation approach.

\subsubsection{Cryptographic Approach}
Secure computation is a cryptographic technique in which functions are evaluated based on a set of distributed inputs without revealing additional information, e.g., the parties’ inputs or intermediate results. Secure multi-party computation, homomorphic encryption, and secret sharing are the most common choices for a secure computing platform. 

Multi-party computation \cite{yao1982protocols} was first introduced to secure the private inputs of multiple participants while they jointly compute an agreed-upon model or function. Formally, $n$ participants $p_1, p_2,…$, and $p_n$ can collaboratively compute $y = f \left( x_1,…,x_n \right) $, where $x_i$ is a secret input that belongs to participants $p_i$, This form of secure computing offers both correctness and privacy. After all, no participant learns anything about the others’ data other than the final result.

Homomorphic encryption \cite{yi2014homomorphic} allows certain mathematical operations, such as addition and multiplication, to be performed directly on ciphertexts. These can then be used as the basis for more complex arbitrary functions. 

\subsubsection{Perturbation Approach}
With privacy concerns in mind, one needs to be cautious of how much information about a participating client is revealed during training. The perturbation approach arises as a natural way of preventing information leaks. By injecting the proper amount of artificial noise into the original data, the statistical information calculated from the perturbed data will be statistically indistinguishable from the original data. 

There are three types of widely used perturbation techniques: differential privacy (DP), additive perturbation, and multiplicative perturbation. DP proposed by Dwork \cite{10.1007/11681878_14}, is the gold standard. The intuition behind DP is to mask the contribution of any individual user by a sufficient level of uncertainty. A randomized mechanism $\mathcal{M}$ is said to be $\left( \epsilon ,\delta \right)$-differentially private if, for any pair of neighboring datasets $\mathcal{D}$ and $\mathcal{D'}$, and for every set of \(S \subseteq Range \left( \mathcal{M} \right) \), if $\mathcal{M}$ satisfies:
\begin{equation}
	{\Pr \left[ {\mathcal{M}\left( \mathcal{D} \right) \in S} \right] \le \exp \left( \varepsilon \right) \cdot \Pr \left[ {\mathcal{M}\left( \mathcal{D'} \right) \in S} \right] + \delta}
\end{equation}
The parameter $\epsilon $ is defined as privacy budget, which measures how alike the two adjacent datasets $\mathcal{D}$ and $\mathcal{D'}$ are to each other. A smaller $\epsilon$ indicates a stronger privacy guarantee.  
If $\delta=0$, then the randomized mechanism $\mathcal{M}$ is degraded into $\epsilon$- DP.

\subsection{Fairness in FL}
\begin{figure}[htp]
	\centering
	\includegraphics[width=8cm]{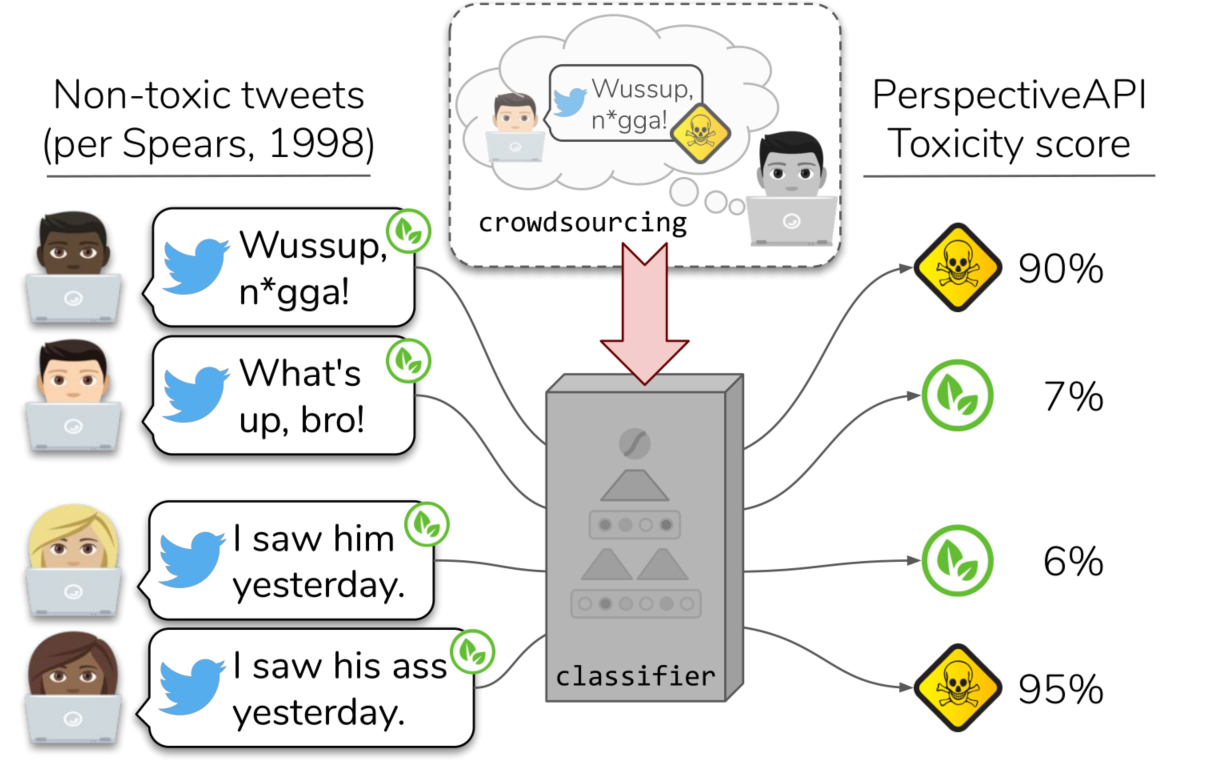}
	\caption{Racial disparities in classifier predictions on tweets written in African-American English and in Standard American English (reproduced from \cite{sap2019risk})}
	\label{fig:bias}
\end{figure}

Algorithms are widely used to assist in making recommendations, assessing loan applications, etc. Several studies have identified the unfairness in different algorithmic scenarios \cite{mehrabi2021survey}. One example is the hate speech detector designed to rate the toxicity score of the given phrases to help companies like Twitter recognize harmful speech. The detector relies on a tool called \textit{Perspective}, which is trained on labeled data. The detector behaves differently towards phrases written in African American English in a racially biased way, as Fig. \ref{fig:bias} shows \cite{sap2019risk}.  

\subsubsection{Bias in FL}
Fairness can be eroded by bias. According to Olteanu et al. \cite{olteanu2019social}, bias can slip into data flows from generation to collection to processing \cite{olteanu2019social}. 
The distributed learning paradigm of FL brings new and unique challenges to our efforts to build fair models. One challenge is that the independent and identical distribution (i.i.d.) assumption no longer holds \cite{kairouz2019advances}. In FL, bias can also be introduced by either the client or the server.

\begin{itemize}
\item \textbf{Client-introduced bias}. Clients can introduce bias in several ways. First, bias exists in a client’s local data – prejudice, underestimation, negative legacies, etc. \cite{kamishima2012fairness}. This bias is then integrated into the global model through client and server interactions. Second, bias can strike when clients are dropped out of the federated setting due to device shutdowns or communication limitations. In these cases, the global model will find it hard to fit the clients’ data properly. The massively distributed data also incurs bias. In this case, the client does not have enough data to capture an underlying distribution, and moreover, the underlying distributions of different clients are probably not the same. \cite{zhao2018federated,Briggs2020FederatedLW,pmlr-v139-li21h} 

\item \textbf{Server-introduced bias}. The server can also add bias. As the coordinator of the learning scheme, the server will sample clients in each round to train a global model with their local data \cite{Nagalapatti_Narayanam_2021}. However, the sampling process is prone to producing bias if it is not done with careful consideration. First, in terms of efficiency, only a fraction of clients are selected in each round \cite{mcmahan2017communication}. Yet the data distribution of only a few selected clients forms an inadequate representation of the actual population distribution. Second, the sampling may be skewed toward certain clients. For instance, to speed up convergence, the server prefers clients that meet specific criteria \cite{nishio2019client,cho2020client}. 
\end{itemize}

\subsubsection{Fairness Notions in FL}
To date, researchers have proposed several definitions of fairness, see, e.g., \cite{10.1145/3194770.3194776,mehrabi2021survey}. These definitions vary from scenario to scenario, and it is unlikely that there will ever be one particular definition of fairness that fits all circumstances. Table \ref{tab:Algorithmic Fairness defintion} lists common algorithmic fairness notions adopted in centralized machine learning. At a high level, two families of definitions exist the \textit{individual} notion and \textit{statistical} notion \cite{chouldechova2018frontiers}.
\begin{itemize}
\item \textbf{Individual notion}. The individual notions ensure fairness between specific pairs of individuals: \textit{"Give similar predictions to similar individuals."} \cite{dwork2012fairness}. Formally, for a set of samples $V$, a distance metric is defined as $d:V\times V\to R$ to measure the similarity. A function $\mathcal{M}: V\to \Delta A$ maps the samples $V$ to the probability distributions over outcomes, and another distance $D$ metric measures the distance between the distributions of outputs. Fairness is achieved if and only if $D\left(\mathcal{M}\left(x\right),\mathcal{M}\left(y\right)\right) \le d\left(x,y\right) $. This family of definitions provides a meaningful guarantee. However, they are at the cost of making significant assumptions, some of which are non-trivial problems in fairness.
	
\item \textbf{Statistical notion}. The statistical notions provide fairness assurance at a statistical level. For the protected demographic groups $G$ (such as racial minorities), some statistical measures are required to be equal across all of these groups. These statistical measures include positive classification rates \cite{6137441,calders2010three,dwork2012fairness}, false positive and false negative rates \cite{hardt2016equality,kleinberg2017inherent}, and positive predictive value \cite{zafar2017fairness,chouldechova2017fair}. Detailed enumeration can be found in \cite{doi:10.1177/0049124118782533,10.1145/3194770.3194776}. This family of definitions requires no assumption over data and can be easily verified. However, statistical notions are insufficient as a fairness constraint, which does not give meaningful guarantees to individuals or structured subgroups of the protected demographic groups. Jiang et al. \cite{jiang2022generalized} generalized the demographic parity \cite{kusner2018counterfactual} to continuous sensitive attribute. 
\end{itemize}
\begin{table}[htp]
\centering
\caption{Definitions of Algorithmic Fairness Notions}
\label{tab:Algorithmic Fairness defintion}
\setlength{\tabcolsep}{2pt} 
\begin{tabular}{ccl}
\hline
\textbf{Fairness notion} & \textbf{Definition} & \multicolumn{1}{c}{\textbf{Explanation}} \\ \hline
\begin{tabular}[c]{@{}c@{}}Individual Fairness\\ \cite{dwork2012fairness}\end{tabular} & $D(M(x),M(y)) \le d(x,y)$ & \begin{tabular}[c]{@{}l@{}}Similar samples receive \\ similar treatment\end{tabular}\\ \hline

\begin{tabular}[c]{@{}c@{}}Eqal Opportunity\\ \cite{hardt2016equality}\end{tabular} & $\Pr[\hat Y =1|A=0, Y=1 ]=\Pr[\hat Y =1|A=1, Y=1]$ & \begin{tabular}[c]{@{}l@{}}Equal true positive rates\\ for protected/unprotected\\ groups\end{tabular} \\ \hline

\begin{tabular}[c]{@{}c@{}}Equal Accuracy\\ \cite{berk2021fairness} \end{tabular} & $\Pr[\hat Y =Y|A=0]=\Pr[\hat Y =Y|A=1] $ & \begin{tabular}[c]{@{}l@{}}Equal prediction accuracy \\ for protected/unprotected \\ groups\end{tabular} \\ \hline

\begin{tabular}[c]{@{}c@{}}Equalized Odds\\ \cite{hardt2016equality} \end{tabular}& \begin{tabular}[c]{@{}c@{}}$\Pr[\hat Y =1|A=1, Y=y ]=\Pr[\hat Y =1|A=0, Y=y],$\\ $y\in \{0,1\}$ \end{tabular} & \begin{tabular}[c]{@{}l@{}}Equal positive rates for\\ protected/unprotected\\ groups\end{tabular} \\ \hline

\begin{tabular}[c]{@{}c@{}}Treatment Equality\\ \cite{berk2021fairness}\end{tabular} & \multicolumn{2}{c}{\begin{tabular}[c]{@{}l@{}}Equal false negatives and fase positives for protected/unprotected groups\end{tabular}} \\ \hline

\begin{tabular}[c]{@{}c@{}}Demographic Parity\\ \cite{kusner2018counterfactual}\end{tabular} & $\Pr[\hat Y |A=0]=\Pr[\hat Y |A=1]$ & \begin{tabular}[c]{@{}l@{}}Outcome is independent \\ of the protected attribute\end{tabular} \\ \hline
\end{tabular}%
\end{table}
Apart from algorithmic fairness, which is measured on sensitive attributes, fairness in FL can also be made from a client's view since clients are naturally grouped by attributes like geographic location, gender and income \cite{du2021fairness}. At a client level, fairness can be evaluated by different metrics.

\begin{definition}[\textit{Good-intent fairness} \cite{mohri2019agnostic}]
The training procedure does not overfit a model to any device at the expense of other clients in FL.
\end{definition}
This metric improves the worst-case performance. Li et al. \cite{li2019fair} took a further step. They tried to ensure a fair FL model for all clients by producing a more uniform model performance across all clients. Fairness is defined as the uniformity of the accuracy distribution across clients in FL. 
\begin{definition}[\textit{Accuracy parity} \cite{li2019fair}]
Consider two trained models, $f(w)$ and $f(\tilde w)$. The model that provides the most uniform performance across all clients will also provide the fairest solution to the FL objective in Eq. (\ref{FL object func}).
\end{definition} 

\subsection{Interactions between Privacy and Fairness} 
Both fairness and privacy are important ethical notions in machine learning and have been extensively studied. However, the majority of current studies in the research community consider fairness and privacy separately. However, the interactions between privacy and fairness are bilateral.
\begin{itemize}
\item \textbf{Privacy degrades fairness}. Several works have observed inconsistent reductions in accuracy caused by private mechanisms on classification \cite{farrand2020neither} and generative tasks \cite{ganev2022robin}. It turns out that privacy mechanisms affect the underrepresented group more than other groups.
\item \textbf{Fairness increases privacy risk}. Fairness comes at the cost of privacy. To ensure fairness, a model is trained to perform equally on data from different groups, even though the underrepresented group didn't have enough data in the training set, which incurs overfit and increases the privacy risk \cite{chang2021privacy}. 
\end{itemize} 

\section{Privacy in FL}
\label{sec: privacy}
With the advent of FL, many claims that user data are now secure. However, even sharing a small fraction of gradients \cite{shokri2015privacy,aono2017privacy} with a server would raise privacy concerns. In FL, there are several unexplored types of privacy attacks: \textit{membership inference attacks}, \textit{property inference attacks}, \textit{model inversion attacks}, and \textit{reconstruction attacks}. This section will outline these attacks before moving on to mitigation techniques and discussions.

\subsection{Membership Inference Attacks}
The adversary’s goal in FL is to determine whether a given sample belongs to a single client’s private training data or of any participants \cite{yang2020federated}. MIAs take occur in different ways in FL.
\subsubsection{White-box and Black-box MIAs}
Based on the access granted to the adversary, MIAs can be divided into black-box and white-box \cite{8835245}. In the black-box setting, the adversary can only obtain a prediction vector computed by the target model while the internal parameters remain secret. MIAs in this setting exploit the statistical differences between a model’s predictions on its training set versus unseen data \cite{shokri2017membership}. Truex et al. \cite{8634878} described a systematic approach to constructing a black-box MIA model and the general formulation of each component in the attack model. 

However, since the global model is shared with all participants for local training, it is often assumed that an adversary has white-box access in FL. The white-box access renders much more information to the adversary, such as the internal parameters of each layer. This enables the adversary to calculate the outputs of each layer. Nasr et al. \cite{8835245} designed a deep learning attack model that separately processes the gradients extracted from different layers of the target model and combines this information to compute the membership probability of a target data point. 

\subsubsection{Training and Inference MIAs}
MIAs can be launched during the training stage or once the model is complete in the inference stage in FL.

In the training stage, the adversary could be the server or any client participating in the training. Both characters have white-box access to the global model and can easily save the snapshots of the global model at each iteration during training. In this way, the adversary obtains multiple versions of the target model over time and acquires the updated information to infer private data \cite{phuong2019privacy}. In addition to passively collecting the updated information, the adversary may further modify the information to allure the victim clients to reveal more information.

In the inference phase, the FL model is well-trained and fixed. The adversary can only perform an inference attack passively. In this case, MIA in FL resembles that in a centralized setting. The attack’s success largely depends on the information that is revealed to the adversary. Melis et al. \cite{melis2019exploiting,} investigated privacy leaks concerning membership during the inference phase and showed that positions of words in a batch could be revealed from a deep learning model.

\subsubsection{Active and Passive MIAs} 
The adversary can conduct MIAs against the FL model actively or passively. For instance, the server can either adaptively modify the aggregate parameters or honestly calculate the global model and passively conduct MIAs \cite{zhang2020gan}. 

Melis et al. \cite{melis2019exploiting} designed MIAs against models operating on non-numerical data (e.g., natural-language text). An embedding layer is equipped for the target model, transforming the inputs into a lower-dimensional vector representation. The adversary passively saves a snapshot of the joint model parameters ${w_t}$. The difference between the consecutive snapshots $\Delta {w _t} = {w_t}-{w_{t-1}}={\Sigma_k}\Delta w _t^k$ reveals the aggregated updates from all participants and hence reveals the membership.

Nasr et al. \cite{8835245} performed active MIAs on FL models by reversing the stochastic gradient descent algorithm and extracting membership information. If the target data point belongs to the training dataset, the attacker’s modifications will be nullified since the target model will descend the model’s gradient for training samples. However, if the target data sample is not used during the training, the target model will not respond to the attacker’s modification. Thus, membership can be deduced. In \cite{hitaj2017deep}, a malicious client actively mislabels the training sample to fool the victim into releasing private information. 
\subsubsection{Insider and Outsider MIAs}
FL involves two types of actors who can access model information: internal actors (participating clients and the server) and external actors (model consumers and eavesdroppers). Therefore, FL systems must withstand potential adversaries within and outside the protocol.

The inner adversary could be a client or a server. Clients are picked at random to participate in a training round. When training with hundreds or millions of clients, malicious clients are highly likely involved, who will attempt to deduce the sensitive information of others \cite{hitaj2017deep}. The real-time nature of FL added to the inner attacker's strength. For example, Zhang et al. \cite{zhang2020gan} trained a GAN as a malicious client during training to infer the data of other clients in FL.
A malicious central server poses a greater threat than a malicious client. Because it can manipulate the global model supplied to victims and obtain more information. Nasr et al. \cite{8835245} launched MIAs from both the client and server sides and witnessed a higher inference accuracy as a curious central server than as a malicious client.

In addition to internal threats, FL also faces potential attacks from adversaries outside the system. Once the FL training is finished and the model is deployed to users, these users may conduct both black- and white-box attacks depending on their access.

\subsubsection{Discussion}
The attacks mentioned above demonstrate the vulnerability of FL to privacy attacks, and these privacy risks stem from two assumptions made within the FL protocols: 1) \textit{The server is trustworthy}. FL gives the server access to each participant's updates in the form of gradients or model parameters containing clients' private information. The server can even purposefully send a modified model to steal information. 2) \textit{Clients are honest}. A malicious client can collect several copies of the global model from the rounds it participates in. In this way, inference phase attacks on data privacy are also plausible during the learning phase. Additionally, adversarial clients may influence and shift the bounds of the model during development rather than just abusing the boundaries of a model's service while it is in production.

\subsection{Property Inference Attacks}
The adversary in property inference attacks attempts to infer the specific property of the subset of the training dataset. The target property may be irrelevant to the classification task (e.g., "wearing glasses" in a gender classification task) and do not characterize the whole class. The attack is made at the population level as opposed to a single sample in MIA. In terms of when the attack is launched, property inference attacks can be classified as \textit{static} or \textit{dynamic} attacks. The static attack is applied after the training phase has concluded and the target training set is fixed. The dynamic attack typically occurs during the training phase in FL. In this instance, the training set is changing dynamically.
\subsubsection{Static Attacks}
The research of property inference attacks dates back to \cite{https://doi.org/10.48550/arxiv.1306.4447}. Ateniese et al. performed a property inference attack against Hidden Markov Models and Support Vector Machine based on a meta-classifier. A set of shadow classifiers were trained on a dataset similar to the target model except for the target property. The meta-classifier is trained with shadow classifiers as the input to find the classifiers trained on the dataset with the target property. Ganju et al. \cite{10.1145/3243734.3243834} extended this attack to a fully connected neural network case. They shared a similar idea, using the gradient of shadow classifiers to train a meta-classifier. Different from \cite{https://doi.org/10.48550/arxiv.1306.4447}, their research focuses on improving the attack efficiency by taking permutation invariance into account.

\subsubsection{Dynamic Attacks}  In every communication round of FL, clients are selected at random. This means the training data is dynamically changing, which weakens the property inference attack because the target property appears unpredictable, thereby diminishing the distinguishability of model updates \cite{wang2022poisoning}. Wang et al. \cite{https://doi.org/10.48550/arxiv.1910.06044} explored property inference attacks within the FL framework. Inspired by the relationship between the changing of neuron weights in the output layer and the sample label, the authors proposed three attacks as an eavesdropper to infer the labels' quantity composition proportion. Recently, Wang et al. \cite{wang2022poisoning} presented a poisoning-assisted property inference attack in FL from the client's viewpoint, aiming at inferring if and when a sensitive property emerges. The authors built their attacks around the realization that regular model updates reflect the shift in data distribution and, in particular. A binary classifier is trained to make predictions based on these periodic model updates. A property-specific poisoning attack is proposed to distort the decision boundary of the shared model on target attribute data. Thus, model updates have a better discerning ability to infer target property.

\subsubsection{Discussion} The MIAs and reconstruction attacks represent two ends of a spectrum of privacy invasion. Property inference attacks locate in the middle and seek to determine if the attackers' target property is present in the training samples. This type of attack is more complex than MIA since the target property doesn't always match the attributes that characterize the classes of the FL model. Nonetheless, a property inference attack poses a greater threat than MIA. Using the real-time nature of FL, the adversary can even infer when the target property appears.

\subsection{Model Inversion Attacks}
Fredrikson et al. \cite{184489} initiated model inversion attacks on tabular data. A subsequent work \cite{fredrikson2015model} extended it to the image data. The attack is formulated as an optimization problem to synthesize the input for a given label: $y:{\max _x}\log {T_y}\left( x \right)$, where ${T_y}\left( x\right)$ is the probability of the model $T$ outputs label $y$ for input $x$. The access could be black-box \cite{fredrikson2015model} or white-box \cite{zhang2020secret,chen2020improved}.

\subsubsection{Black-box Attacks}
In the black-box setting, the attacker can only make prediction queries to the model. Fredrikson et al. \cite{fredrikson2015model} built attack algorithms following the maximum a posterior principle. Their attack recovered a recognizable image of a person given only API access to a facial recognition system and a specific name of a target person. Yang et al. \cite{yang2019adversarial} engineered an inversion model a perform the inversion attacks. The adversary composed an auxiliary set assumed generic enough to retain meaningful information to regularize the ill-posed inversion problem \cite{salem2020updates}. However, the target model is usually assumed to be simple networks, and the generalization to complex models is not trivial. The inversion problem of a neural network is non-convex, and the optimization suffers minimal local problems, which leads to poor attack performance. 

\subsubsection{White-box Attacks} 
n the white-box setting, the attacker has complete knowledge of the model. Zhang et al. \cite{zhang2020secret} sketched a generative model to learn an informative prior from the public dataset. This prior is then used to regulate the inversion problem. Benefiting from this, the authors revealed private training data of DNNs with high fidelity. Chen et al. \cite{chen2020improved} boosted \cite{zhang2020secret}'s methods. They leveraged the target model to label a public dataset, and a GAN model was trained to distinguish not only 
real and synthesized samples but also labels. They also modeled the private data distribution to reconstruct representative data points better. The success of model inversion attack benefits from an informative prior.

\subsubsection{Discussion}  MIAs can be performed with either black-box access or white-box access. When given black-box access, the attack's success heavily relies on the auxiliary dataset, which is assumed to share the same generic features as the private target dataset. Furthermore, the target models in this category are usually simple due to limited access. In the white-box case, the target models extend to DNNs. Most attacks implement GAN to synthesize samples to mimic the private samples regarding the soft labels. This kind of attack is less common compared with reconstruction attacks in FL.

\subsection{Reconstruction Attacks}
Unlike MIAs, reconstruction attacks attempt to retrieve training data and pose a much more severe threat to privacy. As demonstrated by Aono et al. \cite{aono2017privacy}, the gradient of the weights is proportional to that of the bias in the first layer of the model, and their ratio approximates the training input. Geiping et al. \cite{geiping2020inverting} demonstrated that it is possible to faithfully reconstruct images at high resolution given knowledge of the parameter gradients. Such a privacy break is possible even for deep neural networks. Huang et al. \cite{huang2021evaluating} evaluated existing reconstruction attacks and defenses. Gupta et al. \cite{gupta2022recovering} extended this attack to text data and successfully reconstructed single sentences with high fidelity for large batch sizes. To date, we know of two kinds of reconstruction attacks, namely \textit{optimization-based attacks} (Opt-based) and \textit{closed-form attacks}.
\begin{table}[htp]
\centering
\caption{Comparison between two reconstruction attack categories} \small
\setlength{\tabcolsep}{2pt} 
\renewcommand{\arraystretch}{1.25}
{%
\begin{tabular}{ccccccc}
\hline
 & \textbf{\begin{tabular}[c]{@{}c@{}}Theoretical\\  guarantee\end{tabular}} & \textbf{Convergence} &  \textbf{\begin{tabular}[c]{@{}c@{}}Running\\  time\end{tabular}} &  \textbf{\begin{tabular}[c]{@{}c@{}}Recovered\\  image\end{tabular}}& \textbf{Applicability} & \textbf{Insight} \\ \hline
Opt-based & No & Local optimal & Slow & With artifacts & No limitation & No \\ \hline
Closed-form & Yes & / & Fast & Original & Limited & Yes \\ \hline
\end{tabular}%
}
\end{table}

\subsubsection{Optimization-based Attack} 
Raw data can be reconstructed from gradients by solving an optimization problem. Given a machine learning model $f(w)$ and the gradient $g = \frac{1}{b}\sum\nolimits_{j = 1}^b {{\nabla _w }{L_w }\left( {x_j^*,y_j^*} \right)} $ computed on a private batch $\left(x^*, y^*\right) \in \mathbb{R}^{b \times d} \times \mathbb{R}^b$ with bath size $b$. The adversary tries to reconstruct $x \in \mathbb{R}^{b \times d}$ as an approximation of the true data ${x^*}$ by solving the following optimization problem:
\begin{equation}
    \arg \mathop {\min}\limits_x {\mathcal{L}_{grad}}\left(x;w,g \right) + \alpha {\mathcal{R}_{aux}}\left( x \right)
    \label{ReconObj}
\end{equation}
The first part of Eq. \ref{ReconObj} pushes the recovered gradients towards the true gradients $g$, hence deducing a better approximation. The regularization term ${\mathcal{R}_{aux}}\left( x \right)$ is used to incorporate the prior knowledge to further improve reconstruction. 

Zhu and Han \cite{zhu2019deep} proposed \textit{DLG} attack with $l_2$ distance as the reconstruction loss $\mathcal{L}_{grad}$. \textit{DLG} starts with randomly generated dummy samples and then iteratively optimizes the dummy samples and labels until they converge. Finally, \textit{DLG} achieves pixel-wise recovery accuracy for image classification and token-wise recovery accuracy for a masked language model. Zhao et al. \cite{zhao2020idlg} improved Zhu and Han's work \cite{zhu2019deep} on convergence speed and reconstruction fidelity by leveraging the relationship between the ground-truth labels and the signs of the gradients.

The optimization problem Eq. \ref{ReconObj} is often under-determined \cite{jeon2021gradient}, as the information in the gradients $g$ is usually insufficient to recover the training data ${x^*}$. Even when the gradient size is substantially bigger than the input data dimension. As demonstrated by Zhu and Blaschko \cite{zhu2020r}, when the learning model is huge, there may be a pair of separate data sharing the same gradient. 

In response to this, one may introduce prior knowledge as a regularization term to narrow the search space, making it more consistent with the underlying distribution of training data.  Yin et al. \cite{yin2021see} utilized the local batch norm as the regularization term since adjacent pixels in natural photographs are likely to have comparable values. They achieved precise recovery of the high-resolution images on complex datasets, deep networks, and large batch sizes. Hatamizadeh et al. \cite{hatamizadeh2022gradvit} extended \cite{yin2021see}'s approach to vision transformers and discovered that, because of the attention mechanism, vision transformers are substantially more sensitive than previously researched CNNs. In the image domain, total variance is another choice \cite{wang2019beyond,geiping2020inverting}.

Selecting a proper loss function can also contribute to attack efficiency. By combining mean square error and Wasserstein distance \cite{arjovsky2017wasserstein}, Ren et al. \cite{10.1145/3510032} achieved a better reconstruction result than \cite{zhu2019deep,zhao2020idlg} in terms of batch size and reconstruction fidelity. Geiping et al. \cite{geiping2020inverting} adopted cosine distances to better capture the observation that the angle between two data points quantifies the change in prediction. With this method, they rebuilt a single high-resolution image and a series of low-resolution photos with a maximum batch size of 100. Jeon et al. \cite{jeon2021gradient} systematically investigated ways to best utilize gradients, including using them to extract prior information. Wei et al. \cite{wei2020framework} conducted a thorough analysis of how different hyper-parameter setups and settings for attack algorithms influence the effectiveness of the attack and its cost. In an effort to investigate the worst-case attack and evaluate the effectiveness of defense methods for reconstruction attacks, Balunovic et al. \cite{balunovic2021bayesian} formulated the gradient leakage problem in a Bayesian framework and analyzed the condition for a Bayes optimal adversary.
\subsubsection{Closed-form Attack}
In an attempt to provide a theoretical understanding of how and when gradients lead to the remarkable recovery of original data, several studies investigated the possibility of recovering the input of a learnable affine function from gradients \cite{aono2017privacy,qian2020can,fan2020rethinking}. Aono et al. \cite{aono2017privacy} initiated the closed-form attack based on gradients. In certain circumstances, an honest-but-curious server could directly calculate individual data from the gradients uploaded by clients \cite{aono2017privacy,zhu2020r}. 

Consider a fully connected layer $Wx+b=z$ with $l=l(f(x),y)$ as the loss function, where $x$ and $z$ are the input and output vectors, respectively. Private data $x$ can be derived from $l$'s gradients w.r.t $W$ and $b$, i.e.,: ${x^T} = \frac{{\partial l}}{{\partial W}} \oslash \frac{{\partial l}}{{\partial b}}$. Here $\oslash$ denotes entry-wise division. The assumption of a fully connected layer holds for the last prediction layers in many popular architectures. As a result, the prediction modules' input, which is the output of the preceding layers, can be rebuilt. These outputs typically include some information about the training data, making them vulnerable to attackers. In this light, the ability to recover ground truth label information from the gradients of the final fully-connected layer, as stated in Zhao et al. \cite{zhao2020idlg}, is very intriguing. 

Despite its inspiring nature, Aono et al.’s work \cite{aono2017privacy} has some limitations. First, it does not apply to convolutional neural networks due to a mismatch in dimensions. Second, it cannot deal with batch inputs. For the batch input $\left\{x_j\right\}_{j=1}^b$, all derivatives are summed over the batch dimension $b$ and the recovered $\bar x$ is merely proportional to the average of batch inputs $\sum\nolimits_{j = 1}^b {{x_j}}$.

To fix the dimension mismatch issue, Zhu and Blaschko \cite{zhu2020r} converted the convolutional layer into a fully connected layer using circulant matrix representation \cite{golub2013matrix} of the convolutional kernel \cite{fan2020rethinking}. The gradients of each layer were interpreted as the \textit{gradient constraints}. Finally, they recursively reconstructed the layer-wise input. However, their implementation can only recover low-resolution images in settings where the batch size equals 1. For the batch input, the algorithm returns a linear combination of the training data.

To address these difficulties with the batch size, Fowl et al. \cite{fowl2021robbing} suggested making minor but malicious changes to the global model to reconstruct the client’s data from a batch of gradient updates. The key idea is to separate the batch data by some quantity $h$, such as image brightness. To this end, an imprint module is added to the global model, which acts as a filter that separates a batch of samples based on quantity $h$. Qian and Hansen \cite{qian2020can} found bias term in the output layer is the key to the success of reconstruction. A fully-connected neural network requires just one node in one hidden layer for single-input reconstruction. In contrast, mini-batch reconstruction requires that the hidden units exceed the input size. Pan et al. \cite{pan2020Exploring} conducted an analytic investigation of the security boundary of the reconstruction attacks. Given a batch input, the secure/insecure boundary of the reconstruction attack was characterized by the number of Exclusively Activated Neurons (ExANs), where the more ExANs, the more likely the attack's success.
\begin{table}
\caption{Comparison of reconstruction attacks in FL} \small 
\label{tab:Reconstruction attacks} 
\setlength{\tabcolsep}{2pt} 
\renewcommand{\arraystretch}{1.2}
\begin{tabular}{ccccccccc}\hline
\multirow{2}{*}{\textbf{Method}} & \multicolumn{2}{c}{\textbf{Objective function}} & \multirow{2}{*}{\textbf{\begin{tabular}[c]{@{}c@{}}Maximal \\ batch size\end{tabular}}} & \multirow{2}{*}{\textbf{\begin{tabular}[c]{@{}c@{}}Opt-based/\\ Closed-form\end{tabular}}} & \multirow{2}{*}{\textbf{\begin{tabular}[c]{@{}c@{}}Theoretical \\ guarantee\end{tabular}}} & \multirow{2}{*}{\textbf{\begin{tabular}[c]{@{}c@{}}Additional\\  information\end{tabular}}} \\ \cline{2-3}
& $\mathcal{L}_{grad}$ & $\mathcal{R}_{aux}$ &  &  &  &  &  \\ \hline
iDLG \cite{zhao2020idlg} & $l_2$  distance & / & 8 & Opt-based & No & No \\ \hline
DLG \cite{zhu2019deep} & $l_2$  distance & / & 8 & Opt-based & Yes & No \\ \hline
\begin{tabular}[c]{@{}c@{}}Inverting \\gradients\cite{geiping2020inverting} \end{tabular} & \begin{tabular}[c]{@{}c@{}}Cosine \\ similarity\end{tabular} & Total variance & 100 & Opt-based & Yes & \begin{tabular}[c]{@{}c@{}}Local updates;\\ BN statistcs\end{tabular} \\ \hline

\cite{wei2020framework} & $l_2$  distance & \begin{tabular}[c]{@{}c@{}}Label-based \\ regularizer\end{tabular} & 8 & Opt-based & Yes & No  \\ \hline

SAPAG \cite{wang2020sapag} & \begin{tabular}[c]{@{}c@{}}Gaussian\\ kerneal \\ based function\end{tabular} & / & 8 & Opt-based & No & No  \\ \hline

R-GAP \cite{zhu2020r} & \begin{tabular}[c]{@{}c@{}}Recursive \\ gradients\end{tabular} & / & 5 & Closed-form & No & No \\ \hline

\begin{tabular}[c]{@{}c@{}}Theory-\\oriented \cite{pan2020Exploring} \\ \end{tabular}& $l_2$  distance & \begin{tabular}[c]{@{}c@{}}$l_1$ distance of\\feature map \end{tabular} & 32 & Closed-form & Yes & \begin{tabular}[c]{@{}c@{}}Exclusive \\ activated \\ neurons\end{tabular} \\ \hline

\begin{tabular}[c]{@{}c@{}}GradInversion\\ \cite{yin2021see}\end{tabular} & $l_2$  distance & \begin{tabular}[c]{@{}c@{}}Group\\ consistency\end{tabular} & 48 & Opt-based & No & BN statistcs \\ \hline

CAFÉ \cite{jin2021catastrophic} & $l_2$  distance & Total variance & 100 & Opt-based & Yes & Batch indices  \\ \hline

GIAS\&GIM \cite{jeon2021gradient} & \begin{tabular}[c]{@{}c@{}}Negative \\cosine \end{tabular}& \begin{tabular}[c]{@{}c@{}}$l_2$ distance in\\  latent space\end{tabular} & 4 & Opt-based & No & No\\ \hline

\begin{tabular}[c]{@{}c@{}}Imprint \\module \cite{fowl2021robbing}\end{tabular} &\begin{tabular}[c]{@{}c@{}} One-shot\\mechanism\end{tabular}  & / & 16384 & Closed-form & Yes &  CDF\\ \hline

GradViT \cite{hatamizadeh2022gradvit} & $l_2$  distance & \begin{tabular}[c]{@{}c@{}}Image prior; \\ Auxiliary\\ Regularization \end{tabular} & 64 & Opt-based & No & \begin{tabular}[c]{@{}c@{}} Auxiliary\\ networks \end{tabular} \\ \hline
\end{tabular}%
\end{table}

\subsubsection{Discussion}
The closed-form attacks outperform optimization-based attacks in several aspects. First, the closed-form attacks provide a theoretical guarantee of convergence. In contrast optimization-based attacks suffer from the local optimum problem since a non-convex optimization may not always converge to a correct solution \cite{geiping2020inverting}. Further, optimization-based attacks are sensitive to initialization \cite{zhu2019deep}, whereas closed-form attacks do not. Second, the deterministic algorithms run by closed-form attacks are faster than optimization-based attacks. Third, closed-form attacks recover the data more accurately, while optimization-based methods, like GradInversion \cite{yin2021see}, recover data with artifacts. Jin et al. \cite{jin2021catastrophic} made a good comparison between different reconstruction attacks in FL. Table \ref{tab:Reconstruction attacks} summarizes their finding and includes some additional results from this study.

\subsection{Privacy-preserving Techniques}
Privacy-preserving machine learning approaches can be roughly classified as cryptographic approaches and perturbation approaches. Cryptographic approaches enable computation over encrypted data and provide rigorous privacy guarantees in the training process. However, they come at a high computational cost compared to the non-encryption alternatives \cite{xu2021privacy}. This computation overhead limits their application in some learning scenarios, particularly in deep neural networks with huge amounts of parameters. As a result, most state-of-the-art privacy-preserving methods are perturbation-based. The perturbation can be accomplished by adding artifact noise into the dataset, such as DP mechanism \cite{geyer2017differentially,wei2020federated}. Or by representing the raw dataset with a surrogate dataset \cite{wu2021fedcg,scheliga2022precode} or abstracting the dataset via sketch techniques \cite{li2019privacy,haddadpour2020fedsketch}.

\subsubsection{Cryptographic Approaches}
Secure multi-party computation is a sub-field of cryptography that executes calculations on data dispersed among multiple parties in such a way that the computation results are only revealed to the participants \cite{yao1982protocols}. It can take the form of homomorphic encryption (HE) or secret sharing.

As one of the defacto privacy-preserving solutions, homomorphic encryption (HE) provides perfect privacy protection in the face of a malicious server. It allows clients to encrypt their updates in such a way that the server may directly aggregate ciphertexts without divulging anything about the plain text underneath. The downside is that encryption followed by decryption will inevitably impose both a computation and a communications overhead. Phong et al. \cite{Phong2018PrivacyPreservingDL} used \textit{additively homomorphic encryption} to ensure no information was leaked to a malicious server. The encrypted aggregation was formulated as follows:
\begin{equation}
  \bf{E}(W_{global}) :=  \bf{E}(W_{global}) + \bf{E}(-\alpha \cdot G_{local})
\end{equation}
where \textbf{E} is a homomorphic encryption operator that supports addition over ciphertexts, $\bf{G_{local}}$ is the aggregated gradient. The decryption key is public to the clients and private to the server, and thus, the client's information is secured. Due to the additively homomorphic property of \textbf{E}, each client is still able to receive the correct updated model $W_{global}$ via decryption. 
\begin{equation}
   \bf{E}(W_{global}) + \bf{E}(-\alpha \cdot G_{local}) = \bf{E}(W_{global}-\alpha \cdot G_{local})
\end{equation}
However, the amount of data transferred between clients and the server is inflated by two orders of magnitude over the vanilla setting \cite{Phong2018PrivacyPreservingDL}. To reduce the communication load,  Zhang et al. \cite{zhang2019pefl} chose a distributed selective stochastic gradient descent (DSSGD) method in the local training phase to achieve distributed encryption and reduce the computation costs. Zhang et al. \cite{zhang2020batchcrypt} presented a BatchCrypt as a simple batch encryption technique. Clients first quantize their local gradients and then encode a batch of quantized updates into a long integer. As a result, the communication overhead is reduced by up to 101 times.  Jiang et al. \cite{jiang2021flashe} further reduced communication overhead by sending only a sparse subset of local states to the server. 

Another drawback is that all participants share the same private key for decryption since homomorphic operations require all values to be encrypted with the same public kewhich degrades privacy protection in the face of malicious clients. To counter this problem, Park and Lim \cite{park2022privacy} sketched a privacy-preserving FL scheme based on a distributed homomorphic cryptosystem that allows clients to have their own unique private key for the homomorphic encryption scheme.

Secret sharing \cite{shamir1979share} is another kind of cryptographic technique. It splits a secret data $\mathcal{D}$ into $n$ pieces such that the secret $\mathcal{D}$ can be easily reconstructed with at least $k$ pieces. However, any set containing less than $k$ piece reveals no information about $\mathcal{D}$.  It enables the server to aggregate at least a certain number of clients’ updates without disclosing any individual client’s contribution. Bonawitz et al. \cite{bonawitz2017practical,mondal2022scotch} proposed a secure aggregation method for FL based on $t$-out-of-$n$ secret sharing. The key idea is to mask the raw data in a symmetric way. Thus, when aggregated by the server, the introduced noise will be nullified. Liu et al. \cite{liu2020secure} incorporated secure sharing into their federated transfer learning framework to protect privacy. Based on an investigation of how secure aggregation parameters influence communication efficiency, Bonawitz et al. \cite{bonawitz2019federated} used quantization to build a communication-efficient secure aggregation scheme. So et al. \cite{so2021turbo} designed Turbo-Aggregate that leverages additive secret sharing and Lagrange coding to reduce the secure aggregation overhead. Shao et al. \cite{shao2022dres} shared a similar ideal, utilizing Lagrange coding to secretly share private datasets among clients.

Even though FL based on a homomorphic encryption scheme can prevent privacy leaks during training, it remains vulnerable to attacks in the inference stage. The trained model embodies the distribution of training data to a certain extent and the privacy risk still exists. Model inversion attack gives such an example. Given the white-box access to a trained model, Zhang et al. \cite{zhang2020secret} successfully discovered the sensitive features \textit{x} associated with a specific label \textit{y}.

\subsubsection{Perturbation Methods}
Due to its theoretical guarantee of privacy protection and its low computational and communication complexity, the DP technique \cite{dwork2008differential} has emerged as the most popular choice for privacy protection among a variety of options. In differential privacy, a proper amount of noise is added to the raw data \cite{gupta2018distributed}, the model \cite{liu2020survey,geyer2017differentially}, the output \cite{chaudhuri2011differentially}, or the gradients \cite{song2013stochastic} to protect privacy. 

Geyer et al. \cite{geyer2017differentially} applied a differentially private mechanism to the FL scenario, where they approximated the averaging operations with a randomized mechanism that provided client-level privacy. Truex et al. \cite{truex2019hybrid} presented an alternative approach that draws on both DP and secure multi-party computation. The clients and the server communicate through a secure channel. Upon receiving a request from the server, the clients upload their answers following the principles of DP. Xu et al. \cite{10.1145/3308560.3317584} took a different approach. They approximated the objective function of a regression problem via polynomial representation and then added Laplace noise to the polynomial coefficients to protect privacy. Khalili et al. \cite{khalili2021improving} exploited an exponential mechanism \cite{mcsherry2007mechanism} to privately select applicants based on the qualification scores predicted by a pre-trained model. 

One concern with perturbation techniques is the trade-off between privacy, accuracy and convergence. A significant noise perfectly protects privacy at the cost of accuracy and convergence. Conversely, a weak noise is futile to privacy attacks \cite{hitaj2017deep}.  Wei et al. \cite{wei2020federated} conducted a theoretical analysis of the convergence behavior of FL with DP and identified the trade-off between convergence performance and privacy protection levels.

Many studies have been published on ways to deal with this trade-off \cite{fan2020rethinking,zhang2021matrix,shokri2015privacy}. Shokri and Shmatikov \cite{shokri2015privacy} suggested randomly selecting and sharing a small fraction of gradient elements (those with large magnitudes) to reduce privacy loss. Fan et al. \cite{fan2020rethinking} leveraged element-wise adaptive gradient perturbations to defeat reconstruction attacks and maintain high model accuracy. In a similar manner, Wei and Liu \cite{wei2021gradient} used dynamic privacy parameters. Introducing noise with a greater variance at the beginning of training and progressively decreasing the amount of noise and variance as training progresses.

Huang et al. \cite{huang2020instahide} proposed \textit{InstaHide}, a combination of cryptographic and perturbation approaches to provide rigorous privacy protection at the cost of minor effects on accuracy.  \textit{InstaHide} encrypts the raw image by mixing it with multiple random images from a large public dataset. After that, it randomly flips the signs of the pixels before using it to train the model.

Yang created et al. \cite{yang2021gain} NISS to avoid the trade-off between accuracy and privacy by permitting clients to collaborate on reducing the total amount of injected noise. In particular, each client's noise is neutralized and distributed to other clients. Theoretically, if all clients are trustworthy, the locally introduced noise can be perfectly offset by the server's aggregation, completely avoiding the privacy accuracy trade-off. A similar idea can be found in Yang et al. \cite{yang2020accuracy}.

\subsubsection{Trusted Execution Environment}
Some researchers use Trusted Execution Contexts (TEEs) like Intel SGX and ARM TrustZone to secure ML training in untrusted environments \cite{tramer2018slalom,mo2020darknetz,gu2019yerbabuena}. With hardware and software safeguards, TEEs secure critical code from other programs. Compared with purely cryptography methods, TEEs provide much better performance since it only requires extra operations to create the trusted environment and communicate between trusted and untrusted components. Gu et al. \cite{gu2019yerbabuena} partitioned DNN models and solely encased the first layers in an SGX-powered TEE to protect input information. Hynes et al. \cite{hynes2018efficient} investigated speeding up the training using Graphics Processing Units (GPU). Tramer et al.  \cite{tramer2018slalom} shared the same concept and offered effective privacy-preserving neural network inference utilizing trusted hardware that delegated matrix multiplication to an untrusted GPU. However, this work does not translate well to FL due to the possible adversary server and limited computation power of the client device. To remedy this, Mo et al. \cite{mo2020darknetz} advocated using the TEE of client devices in tandem with model partitioning to defend against MIA. The model is divided into two halves, and the final layers are calculated within TEE. Kato et al. \cite{kato2022olive} proposed to combine DP with TEE in FL in the presence of an untrusted server. The models are aggregated within the TEE of the server's device. \cite{chen2020training}

\subsubsection{Discussion}
Table \ref{table:defense} summarized and compared the existing defense techniques. Cryptographic approaches preserve privacy to a great extent while suffering from computational complexity and are less feasible. The perturbation approaches trade-off privacy for model performance. Several inspiring works demonstrate that it may be possible to avoid that trade-off through either client collaborations to neutralize locally added noise on the server side or by using a surrogate dataset to protect the raw data without adding noise. The cryptographic approaches only ensure that no information will leak during training. They do not protect privacy during the inference stage. In contrast, the perturbation approaches (e.g., DP) protect privacy in both the training and inference stages. One may combine cryptographic and perturbation approaches to obtain better privacy protection throughout the machine learning pipeline. 

\begin{table}[htp]
    \centering
    \caption{Privacy-preserving methods in FL}
    \label{table:defense} \small
    \resizebox{\textwidth}{!}{
    \setlength{\tabcolsep}{2pt}
    \begin{tabular}{|c|c|l|l|l|}
    \hline
    \textbf{Attack} & \begin{tabular}[c]{@{}l@{}}\textbf{Defense}\\ \textbf{method} \end{tabular}&  \multicolumn{1}{c|}{\textbf{Rationale}} & \multicolumn{1}{c|}{\textbf{Advantage}} & 
    \multicolumn{1}{c|}{\textbf{Disadvantage}} \\ \hline
    
    \multirow{10}{*}{RA} & \begin{tabular}[c]{@{}c@{}}HE\\ \cite{aono2017privacy,shin2021homomorphic,jiang2021flashe} \end{tabular}& \begin{tabular}[c]{@{}l@{}} Gradients are encrypted\end{tabular} &\multicolumn{1}{c|}{Accurate}  &\begin{tabular}[c]{@{}l@{}} 1. Vulnerable if there are\\ multiple colluding entities; \\2. Ineffective at inference \end{tabular}\\ \cline{2-5}
    
     & \begin{tabular}[c]{@{}c@{}} Secret sharing\\ \cite{bonawitz2017practical,mondal2022scotch,shao2022dres}  \end{tabular} &  \begin{tabular}[c]{@{}l@{}} Hiding information about \\clients’ individual update, \\except for their sum \end{tabular} & \begin{tabular}[c]{@{}l@{}}1. Accurate; 2. Robust to \\ users dropping out \end{tabular}&\begin{tabular}[c]{@{}l@{}} Ineffective at inference \end{tabular}\\ \cline{2-5}
        
     & \begin{tabular}[c]{@{}c@{}}Variational\\ bottleneck\\ \cite{scheliga2022precode}\end{tabular} &  \begin{tabular}[c]{@{}l@{}} Using surrogate gradient \\to protect privacy. \end{tabular}& \begin{tabular}[c]{@{}l@{}} Keep training process\\ and performance intact \end{tabular}& \begin{tabular}[c]{@{}l@{}} Limit to optimization-based \\attack\end{tabular} \\ \cline{2-5}
        
     & \begin{tabular}[c]{@{}c@{}} Gradient\\ compression \\ \cite{https://doi.org/10.48550/arxiv.2011.04926, zhu2019deep,li2022soteriafl} \end{tabular} &  \begin{tabular}[c]{@{}l@{}} Compressing gradients to \\prevent reconstruct private\\  data by matching gradients \end{tabular} & \begin{tabular}[c]{@{}l@{}} 1. Easy to implement;\\ 2. Reduce communication\end{tabular} & \multirow{4}{*}{\begin{tabular}[c]{@{}l@{}} Requires considerable noise,\\ degrades model performance,\\  and increases convergence\\ time \end{tabular}}\\ \cline{1-4}

    \multirow{4}{*}{\begin{tabular}[c]{@{}c@{}} RA,\\ MIA,\\ and PIA \end{tabular}} & \begin{tabular}[c]{@{}c@{}}DP\\\cite{wei2020framework,chen2020differential,ma2021nosnoop}\end{tabular} &  \begin{tabular}[c]{@{}l@{}} Hiding private information \\by injecting noise to the\\ raw data, model, or output\end{tabular} & \begin{tabular}[c]{@{}l@{}} 1. Easy to implement;\\ 2. Long-term protection\end{tabular}& \\ \cline{2-5}

     & \begin{tabular}[c]{@{}c@{}}TEEs\\ \cite{mo2020darknetz,kato2022olive} \end{tabular}&  \begin{tabular}[c]{@{}l@{}} Isolating part of networks\\ from the untrusted \\environments \end{tabular} &  
     \multicolumn{1}{c|}{Reduce computation} & \multicolumn{1}{c|}{Limited memory space}\\ \hline
    \multicolumn{5}{l}{\small RA: Reconstruction attack; MIA: Membership inference attack; PIA: Property inference attack.}
    \end{tabular}
    }
\end{table}

\begin{table}[htp]
\caption{Comparison of privacy attacks in FL} \small 
\label{tab:different privacy attacks} 
\setlength{\tabcolsep}{2pt} 
\begin{tabular}{|c|c|c|c|l|l|}
\hline
\textbf{Attack} & \textbf{Ref.} & \textbf{Access} & \begin{tabular}[c]{@{}c@{}}\textbf{Attack} \\ \textbf{interface} \end{tabular}  & \multicolumn{1}{c|}{\textbf{Assumptions}}  &  \multicolumn{1}{c|}{\textbf{Key technique}} \\ \hline

\multirow{2}{*}{\rotatebox[origin=c]{90}{\begin{tabular}[c]{@{}c@{}}Membership \\ Inference Attack\end{tabular} }}& \cite{shokri2017membership} & BB & \begin{tabular}[c]{@{}c@{}} Confidence \\vector \end{tabular} & \begin{tabular}[l]{@{}l@{}}Knowledge about population\\ data \end{tabular}& \begin{tabular}[l]{@{}l@{}}Inferring from the discrepancies of \\predictions on training set versus\\  unseen data\end{tabular} \\ \cline{2-6} 

 & \cite{8835245} & WB & \begin{tabular}[c]{@{}c@{}} Model\\parameters  \end{tabular} & \begin{tabular}[c]{@{}l@{}}A significant portion of the\\ training data\end{tabular} & \begin{tabular}[l]{@{}l@{}}Reversing the SGD algorithm \end{tabular} \\ \cline{2-6} 
 
 & \cite{leino2020stolen} & WB & \begin{tabular}[c]{@{}c@{}} Model\\parameters  \end{tabular} & \begin{tabular}[l]{@{}l@{}}A proxy dataset sampled from \\the ground-truth distribution \end{tabular} & \begin{tabular}[c]{@{}l@{}}Inferring from parameter differences  \\between the target and proxy model\end{tabular} \\ \hline
 
 \multirow{1}{*}{\rotatebox[origin=c]{90}{\begin{tabular}[c]{@{}c@{}}Property \\ Inference Attack\end{tabular} }} &\cite{https://doi.org/10.48550/arxiv.1306.4447} & WB & \begin{tabular}[c]{@{}c@{}} Model\\parameters  \end{tabular} & \begin{tabular}[c]{@{}l@{}} 1.Knowledge about training \\data structure; 2. Access to the \\ground-truth distribution \end{tabular} & \begin{tabular}[c]{@{}l@{}}  A meta-classifier to infer properties \\ from multiple shadow classifier\\ parameters \end{tabular}\\ \cline{2-6} 

&\cite{wang2022poisoning} & WB & \begin{tabular}[c]{@{}c@{}} Model\\updates \end{tabular} & \begin{tabular}[c]{@{}l@{}} Adversary can manipulate\\ more than one device in FL \end{tabular} & \begin{tabular}[c]{@{}l@{}} Inferring other clients' data from the\\ periodic model updates \end{tabular} \\ \cline{2-6}

&\cite{https://doi.org/10.48550/arxiv.1910.06044} & WB & \begin{tabular}[c]{@{}c@{}} Model\\updates \end{tabular} & \begin{tabular}[c]{@{}l@{}} 1. Client's average label count;\\2. Number of samples per label\end{tabular} & \begin{tabular}[c]{@{}l@{}} 
Inferring from the layer neuron \\weight changes \end{tabular} \\\hline 

\multirow{6}{*}{\rotatebox[origin=c]{90}{Model Inversion  Attack}} & \cite{fredrikson2015model} & BB/WB & \begin{tabular}[c]{@{}c@{}}Confidence \\vectors /\\ Model \\parameters \end{tabular} & \begin{tabular}[l]{@{}l@{}} 1. Side information; 2. Simple \\ networks \end{tabular} & \begin{tabular}[c]{@{}l@{}}Optimizing the input to maximize\\ confidence vectors subject to the\\ classification matches the target\end{tabular} \\ \cline{2-6}

& \cite{yang2019adversarial} & BB & \begin{tabular}[c]{@{}c@{}} Confidence \\vectors\end{tabular} &\begin{tabular}[c]{@{}l@{}} 1. Auxiliary dataset retains \\ meaningful prior information; \\2. Simple networks\end{tabular} & \begin{tabular}[c]{@{}l@{}} An inverse model approximates the\\ mapping between predictions \\and images\end{tabular} \\ \cline{2-6} 

& \cite{zhang2020secret} & WB & \begin{tabular}[c]{@{}l@{}}Feature \\extractor \end{tabular}  &\begin{tabular}[c]{@{}l@{}} An auxiliary dataset retains\\ meaningful  prior information\end{tabular} & \begin{tabular}[c]{@{}l@{}} 1. Distilling prior knowledge from \\an auxiliary dataset via GAN;\\ 2. Optimizing the generate image to\\ maximize likelihood\end{tabular} \\ \hline

\multirow{13}{*}{\rotatebox[origin=c]{90}{Reconstruction Attack}} & \cite{wang2019beyond} & WB & \begin{tabular}[c]{@{}c@{}}Client \\ updates \end{tabular} & \begin{tabular}[l]{@{}l@{}} 1. Shallow target model; \\ 2.  Low-resolution image \end{tabular} & \begin{tabular}[c]{@{}l@{}}A GAN with multi-task  discriminator\\ to enhance fidelity and identify client \end{tabular} \\ \cline{2-6}

&\cite{geiping2020inverting}& WB & \begin{tabular}[c]{@{}c@{}}Client\\ updates \end{tabular} & \begin{tabular}[l]{@{}l@{}} 1. Honest-but-curious server in\\ FL; 2. Small batches \end{tabular} & \begin{tabular}[l]{@{}l@{}} Optimizing the image to get a similar\\ change in model prediction as the \\target images\end{tabular} \\ \cline{2-6}

& \cite{zhu2019deep} & WB & Gradients & \begin{tabular}[l]{@{}l@{}}1. Small image size; 2. Single\\ batches; 3. Target model is\\twice differentiable \end{tabular} & \begin{tabular}[l]{@{}l@{}}Jointly optimizing inputs and labels\\ to match gradients\end{tabular} \\ \cline{2-6}

&\cite{hatamizadeh2022gradvit} & WB & Gradients & \begin{tabular}[l]{@{}l@{}}Auxiliary networks provide\\ image  prior\end{tabular} & \begin{tabular}[l]{@{}l@{}}Optimizing inputs to match  the target\\ gradients with image prior and \\ auxiliary regularizer \end{tabular} \\ \cline{2-6}

& \cite{yin2021see} & WB & BN layers & \begin{tabular}[l]{@{}l@{}}1. BN layers in target model;\\ 2. No repeating labels in a \\batch. \end{tabular} & \begin{tabular}[l]{@{}l@{}} Optimizing the input to match the\\ statistics of BN layers of the target\\ model\end{tabular} \\ \hline
\multicolumn{6}{l}{\small BB: black-box; WB: white-box}
\end{tabular}%
\end{table}

\subsection{Discussion of privacy attacks and defenses in FL}
This section reviews existing privacy attacks and defense approaches in FL. Table \ref{tab:different privacy attacks} summarized the existing privacy attacks in FL. From the attacker's perspective, FL differentiates from the centralized counterparts in sever aspects: 1) \textit{The active attacker in FL.} Due to the collaboration between clients and the server, an adversary could actively attack for victim's private data. For example, the attacker may maliciously reverse the gradients \cite{8835245} or mislabel the training sample \cite{hitaj2017deep} to neutralize the benign clients' efforts and fool them into revealing more information about their private data. Hence making them more venerable compared to centralized machine learning. 2) \textit{The real-time nature of FL strengthens the attacker's ability.} During the training process, the adversary could adaptive change their strategy to infer the victim's private data. As a result, the adversary can even infer a specific clients' data \cite{wang2019beyond} when the target features appear in FL \cite{https://doi.org/10.48550/arxiv.1910.06044,wang2022poisoning}, which is way more severe than centralized setting. 3) \textit{Gradients are shared between clients and server.} Unlike the centralized counterpart, where the adversary could at most access the white-box access to the target model. In FL, gradients are repeatedly shared between clients and the server. Which enables gradient-based privacy attacks. As shown by \cite{zhu2019deep, yin2021see}, the malicious server can reconstruct clients' training data at the pixel level by minimizing the distance to the target gradients.

From the defender's perspective, protecting privacy in FL is also different from that in the centralized scenario. 1) \textit{Malicious could be the server or any client}. FL allows clients to keep private data local. A central server is designed to orchestrate the training process. Which complicated privacy protection. The adversary could be the server \cite{wang2019beyond,10.1145/3510032,geiping2020inverting} or the client \cite{hitaj2017deep}. The malicious adversary is able to infer the target client's privacy passively or actively. For example, sending the modified global model to the target client to probe private data \cite{fowl2021robbing}. This brings challenges to defending against potential privacy attacks. DP is a prevailing choice, but it degrades performance. The cryptographic approaches, like HE, and MPC, retain both privacy and performance at the cost of computation overhead, which is a more severe issue in FL since most clients' devices are limited in computation power. 2) \textit{Training and inference stage privacy attacks.} Different from centralized machine learning, where the major privacy leakage happens at the inference stage, i.e., malicious users probe private training data by inferring the target model. In FL, the attacks could happen during or after the training. This requires the defenders to be aware of both possibilities. The cryptographic approaches proved provable privacy protection during the training stage. However, fail at the inference stage since training distribution is embedded in the trained model's parameters. The perturbation approaches, e.g., DP, provides long-term protection and covers both the training and inference stage. One can hide sensitive information from adversaries by adding appropriate noise to the training data.

\section{Fairness in FL}
\label{sec: fairness}
Fairness, as discussed in the centralized setting, is mainly defined at either the group level \cite{kusner2018counterfactual,hardt2016equality} or the individual level \cite{dwork2012fairness}. In the FL scenario, fairness has a broader definition. Beyond the long-established \textit{algorithmic fairness} \cite{kusner2018counterfactual,hardt2016equality}, \textit{client-level fairness} \cite{yu2020fairness,li2019fair,mohri2019agnostic,pmlr-v119-martinez20a} arises as a new challenge in FL. 

\subsection{Algorithmic Fairness}
Algorithmic fairness is commonly used to describe the discrepancies in algorithm decisions made across distinct groups as defined by a sensitive attribute. FL often involves a deep neural network with redundant parameters and is pruned to overfit the privileged groups. Various debiasing methods have been devised for different applications, including machine learning \cite{zafar2017fairness,zhang2019fairness,berk2017convex,bolukbasi2016man,brunet2019understanding}, representation learning \cite{louizos2016variational,zhang2018mitigating}, and natural language processing \cite{zhao2018gender,may2019measuring,bordia2019identifying,font2019equalizing}. These methods vary in detail but share similar principles. Following the data flow, debiasing methods can be grouped into \textit{pre-processing, in-processing} and \textit{post-processing} categories, which address the discriminate issues at three distinct stages of the data’s handling \cite{d2017conscientious}. 
\subsubsection{Pre-processing} 
Pre-processing tries to remove the underlying discrimination from the data typically by 1) altering the values of the sensitive attributes/class labels; 2) mapping the training data to a new space where the sensitive attributes and class labels are no longer relevant \cite{feldman2015certifying,kamiran2012data,kamiran2010classification}; or 3) reweighting the samples in the training dataset to compensate for skewed treatment \cite{kamiran2012data}. 

Intuitively, by training a classifier on discrimination-free data, it is likely that the resulting predictions will be discrimination-free. Inspired by this idea, Kamiran and Calders \cite{kamiran2012data} proposed three types of pre-processing solutions to learn a fair classification, \textit{messaging, reweighing and sampling}. Feldman et al. \cite{feldman2015certifying} investigated the problem of identifying and removing disparate impacts in the data. Xu et al. \cite{xu2018fairgan} proposed FairGAN, which generates fair data from the original training data and uses the generated data to train the model. Abay et al. \cite{abay2020mitigating} proposed two reweighting methods for the FL setting \cite{kamiran2012data}, \textit{local reweighing} and \textit{global reweighing with DP}. 

Notably, these pre-processing techniques require access to the training data, which violates the privacy principles of FL. As a result, these types of techniques can only be deployed locally on each client. However, in the presence of data heterogeneous among clients, local debiasing cannot provide fair performance for an entire population \cite{cui2021addressing}. 

\subsubsection{In-processing}
In-processing modifies traditional learning algorithms to address discrimination \cite{zafar2017fairness,goh2016satisfying,kamishima2012fairness,berk2017convex,10.1145/3278721.3278779,zemel2013learning}. Such as adding a regularization term to the loss function. Berk et al. \cite{berk2017convex}, for example, incorporated a family of fairness regularizers into the objective function for regression problems. These regularizers span the range from notions of group fairness to individual fairness. They also create a trade-off between accuracy and fairness. 

Another in-processing option is imposing constraints. Zhang et al. \cite{10.1145/3278721.3278779} used a GAN to constrain the bias in a model trained on biased data. During training, the scheme simultaneously tries to maximize the accuracy of the predictor while minimizing the ability of the adversary to predict the protected variable. In FL, G’alvez et al. \cite{Galvez2021EnforcingFI} studied the notion of group fairness as an optimization problem with fairness constraints. Papadaki et al.  \cite{papadaki2021federating} formulated a min-max optimization problem to investigate group fairness in scenarios where population data were distributed across clients. Ezzeldin et al. \cite{ezzeldin2021fairfed} replaced the aggregation protocol FedAvg with FairFed, which adaptively updates the aggregating weights in each round to improve group fairness. Clients whose local measurements match the global fairness measure are given preferential treatment. Khedr et al. \cite{khedr2022certifair} add a regularizer term to minimize the average loss in fairness across all training data.
\subsubsection{Post-processing}
Post-processing addresses discrimination issues after the model is trained and doesn't need to change the training process. The general methodology of post-processing algorithms is to take a subset of samples and change their predicted labels to meet a group fairness requirement \cite{calders2010three,hardt2016equality,bolukbasi2016man,NIPS2017_b8b9c74a,canetti2019soft,8682620}. 

Hardt et al. \cite{hardt2016equality} proposed a post-processing technique to construct a non-discriminating predictor $\tilde Y$ from a learned discriminatory binary predictor $\hat{Y}$. Only access to the prediction $\hat{Y}$, the protected attribute $A$ and target label $Y$ in the data are required, while details of the mapping of features $X$ to prediction $\hat{Y}$ are not needed.  Canetti et al. \cite{canetti2019soft} and Pleiss et al. \cite{NIPS2017_b8b9c74a} shared the key characteristics as Hardt et al.’s \cite{hardt2016equality} work. Lohia et al. \cite{luo2021no} designed a post-processing method to increase both individual and group fairness. Salvador et al. \cite{salvador2021faircal} introduced a conditional calibration method for fair face verification. Their method clusters images into different sets and assigns distinct thresholds to different sets.

\subsubsection{Discussion}
Three different kinds of debiasing methods are at hand in centralized machine learning. However, solutions in the centralized setting cannot be applied directly in the FL scenario due to limitations with the training data. More specifically, in federated settings, the clients usually have limited amounts of data. Hence, a single client can't accurately represent the true distribution over all clients. Consequently, debiasing data before training is not an option. Another limitation is that direct access to local data is prohibited on the server side. Nevertheless, canny researchers have found inspiration from and workarounds to these issues. Gálvez et al. \cite{Galvez2021EnforcingFI}, for example, bypassed this access restriction by using statistics to guide the model’s training instead of the raw data.

\subsection{Client Fairness}
Client fairness in FL is another different fairness notion than algorithmic notions. Ideally, the models produced from FL should capture clients' data distributions and generalize well when deployed on the client side. However, data distribution usually varies among clients. As a result, the global model has inconsistent performance on different clients' dataset. At the client level, a FL protocol is considered to be fair if the performance fluctuates within a limited range, i.e., the variance in the model’s performance across clients falls under a predefined threshold. To this end, two lines of research exist to mitigate fairness issues in FL. These are the \textit{single model approach} and the \textit{personalized models approach}.
\subsubsection{Single Model Approach} 
The single model approach trains a single global model for all clients as a standard FL scheme. Here, the focus is on solving any statistical heterogeneity during the training phase rather than smoothing the distribution difference. 
\begin{itemize}
\item \textbf{Data augmentation} is a straightforward solution to statistical heterogeneity. It increases data diversity on the client side. Several researchers have studied ways to enhance the statistical homogeneity of local data in FL \cite{zhao2018federated,jeong2018communicationefficient,Hao_2021_CVPR}. Zhao et al. \cite{zhao2018federated} suggested a data share scheme, which creates a globally-shared dataset that is balanced by class. The experiment shows a 30\% improvement on accuracy with only 5\% globally shared data. Jeong et al. \cite{jeong2018communicationefficient} proposed \textit{FAug}. Clients first collectively train a GAN model, which is then distributed to clients to augment their local data towards yielding an i.i.d dataset. 

\item \textbf{Client Selection} is another strategy that focuses on sampling data from a homogeneous distribution. Wang et al. \cite{9155494} proposed a control framework to actively select the best subset of clients in each training round. In Yang et al.’s \cite{yang2020federated} method, the local data distribution is estimated first by comparing local updated gradients and gradients inferred from a balanced proxy dataset. The client selection algorithm based on a combinatorial multi-armed bandit was designed to minimize the effect of class imbalances. 

\item \textbf{Agnostic approach} trains a robust model against a possible unknown testing distribution. Mohri et al. \cite{mohri2019agnostic} modeled testing distributions as an unknown mixture of all $m$ clients' data. The global model is optimized for all possible target distributions. This makes the global model more robust to an unknown testing distribution. Du et al. \cite{du2021fairness} introduced a fairness constraint into Mohri et al.'s method \cite{mohri2019agnostic} and proposed \textit{AgnosticFair}, a fairness-aware FL framework. Their method can provide both \emph{Good-intent fairness} and \emph{demographic parity}

\item \textbf{Reweighting} tries to train a fair model by assigning suitable aggregating weights $p_k$ in Eq. \ref{FL object func} to clients. Inspired by $\alpha$-fairness notions \cite{lan2010axiomatic,mo2000fair}, Li et al. \cite{li2019fair} sketched \textit{$q$-Fair FL} ($q$-FFL) to foster fairer accuracy distribution across all clients by up-weighing clients with lower performance during aggregation. Huang et al. \cite{huang2020fairness} shared a similar idea where, for each round of aggregation, clients with lower accuracy or less training participant times are assigned higher aggregation weights. 
\end{itemize}

\subsubsection{Personalized Models Approach} 
Instead of smoothing the statistical heterogeneity, in \textit{personalized FL}, multiple distinct models are trained for clients with different data distributions. A global model is first trained collaboratively and then personalized to clients using private data. In this way, clients can benefit from other clients’ data and solve the issue of statistical heterogeneity. Mansour et al. \cite{mansour2020three} designed and analyzed three approaches to learning personalized models to learn personalized models. Kulkarni et al. \cite{kulkarni2020survey} conducted a brief overview of personalized FL. Chen et al. \cite{chen2022pfl} provided a comprehensive benchmark of various personalized FL methods. Tan et al. \cite{tan2021towards} systematically reviewed this topic and classified personalized FL techniques in terms of data-based and model-based approaches. Here, we summarize their conclusions.
\begin{itemize}
\item \textbf{Multi-task learning} treats building models for each client as different tasks. Smith et al. \cite{smith2017federated} pioneered this approach and explored personalized FL via a multi-task learning framework. \cite{agarwal2020federated,liang2020think} followed this principle. Dinh et al. \cite{dinh2021fedu} proposed FedU, which incorporates a Laplacian regularization term into the optimization problem to leverage relationships between clients.

\item \textbf{Model interpolation} trains local and global models simultaneously, where the global model is used for its generalization ability, and the local model is used to improve local performance. Hanzely and Richtárik \cite{hanzely2021federated} formulated an optimization problem that learns a mixture of the global and local models. The local model is trained solely on each client’s private data. Softly-enforced similarity from multi-task learning is borrowed to discourage the local model from departing too much from the mean model. Deng et al. \cite{deng2020adaptive} and Mansour et al. \cite{mansour2020three} adopt a similar formulation to determine the optimal interpolation of the local and global models. In Zhang et al.'s  \cite{zhang2020personalized} work, clients are given access to multiple models uploaded by other clients to evaluate how much they will benefit from these models. An optimal combination is then used as a personal update. Lin et al. \cite{lin2022personalized} investigated the trade-offs between local and global models.

\item \textbf{Parameter decoupling} learns local parameters as an independent task performed locally. The local model is designed to assist in personalizing the global model to local distributions. Liang et al. \cite{liang2020think} devised the local-global federated averaging algorithm, which jointly learns compact local representations for each client and a global model across all devices. Chen and Chao \cite{chen2021bridging} decomposed a FL model as a generic predictor, which is trained globally, along with a personalized predictor that is trained locally. The personalized predictor is formulated as a lightweight, adaptive module on top of the generic predictor.

\item \textbf{Transfer learning} is a practical training paradigm that leverages knowledge from a source domain to help train a model in a target domain. The performance of transfer learning depends on the similarity between the two domains. Federated transfer learning was first introduced by Liu et al. \cite{liu2020secure}. Since clients in the same federation usually share the same domain, a FL scheme would make a suitable partner for transfer learning. Li and Wang \cite{li2019fedmd} subsequently proposed FedMD, which combines transfer learning and knowledge distillation. Each client performs transfer learning by training a model to converge on a public dataset and subsequently fine-tune it on local data. 

\item \textbf{Clustering} arranges clients into different groups and trains a specific model for each group. Ghosh et al. \cite{ghosh2020efficient} iteratively determines the membership of each client to a cluster and optimizes each of the cluster models via gradient descent in a distributed setting. Sattler et al. \cite{sattler2020clustered} clusters clients according to the cosine similarity between the clients’ gradient updates. This allows clients with a similar distribution to profit from one another while minimizing detrimental interference from others. In Briggs et al.’s \cite{Briggs2020FederatedLW} method, a clustering step is periodically inserted into the training process to cluster clients based on their local updates. The clusters are then trained individually and in parallel on specialized models. Mansour et al. \cite{mansour2020three} proposed hypothesis-based clustering, partitioning clients into $q$ clusters and finding the best hypothesis for each cluster.
 
\item \textbf{Regularization} prevents overfitting when training models and has been used in several studies to remedy the weight divergence problem in FL settings. Li et al. \cite{li2020federated} introduced a proximal term that considers the differences between global and local models to limit the effect of local updates. Yao et al. \cite{yao2020continual} considered parameter importance in the regularised local loss function by using elastic weight consolidation \cite{kirkpatrick2017overcoming}. In addition, a regularization term is introduced to penalize the deviation of the local model from the global model.

\item \textbf{Meta-learning} aims to leverage prior experience with other tasks to facilitate the learning process. The resulting models are highly-adaptable to new heterogeneous tasks \cite{nichol2018firstorder,finn2017model}. Fallah et al. \cite{NEURIPS2020_24389bfe} studied a personalized variant of FedAvg based on model-agnostic meta-learning formulation. The proposed Per-FedAvg algorithm looks for an initial model that performs well after one step of the local gradient update on each client's data. Others have interpreted FedAvg as a meta-learning algorithm, breaking it into two stages of training and fine-tuning to optimize personalized performance and model convergence \cite{jiang2019improving,khodak2019adaptive}.   
\end{itemize}
\begin{table}[]
\renewcommand{\arraystretch}{1.2}
\caption{Summary of Fairness-aware FL}
\label{tab: FAFL} \small
\begin{tabular}{cccccc}
\toprule
\textbf{Reference} & \textbf{\begin{tabular}[c]{@{}c@{}}Single\\ Model\end{tabular}} & \textbf{\begin{tabular}[c]{@{}c@{}}Personalized\\ Model\end{tabular}}  & \textbf{\begin{tabular}[c]{@{}c@{}}Algorithmic \\ Fairness\end{tabular}} & \textbf{\begin{tabular}[c]{@{}c@{}}Client \\ Fairness\end{tabular}} & \textbf{Method}\\ 
\midrule
\cite{zhao2018federated,jeong2018communicationefficient,Hao_2021_CVPR} & \checkmark &   &  & \checkmark & Data Augmentation\\
\cite{9155494,9141436,yang2020federated} & \checkmark &   &  & \checkmark & Client Selection\\
\cite{papadaki2021federating} & \checkmark &   & \checkmark &  & Agnostic approach\\
\cite{du2021fairness} & \checkmark &   & \checkmark & \checkmark & Agnostic approach\\
\cite{mohri2019agnostic,hu2020fedmgda} & \checkmark &   &  & \checkmark & Agnostic approach\\
\cite{ezzeldin2021fairfed} & \checkmark &   & \checkmark &  & Reweight\\
\cite{li2019fair,huang2020fairness} & \checkmark &   &  & \checkmark & Reweight\\
\cite{10.1145/3493700.3493750,Galvez2021EnforcingFI,yue2021gifairfl} & \checkmark &   & \checkmark &  & Regularization\\
\cite{mansour2020three,ghosh2020efficient,sattler2020clustered,Briggs2020FederatedLW} &  & \checkmark  &  & \checkmark & Cluster\\
\cite{mansour2020three,hanzely2021federated,deng2020adaptive,zhang2020personalized} &  & \checkmark  &  & \checkmark & Model interpolation\\
\cite{smith2017federated,agarwal2020federated,liang2020think,dinh2021fedu} &  & \checkmark  &  & \checkmark & Multi-task learning\\
\cite{liang2020think,chen2021bridging} &  & \checkmark  &  & \checkmark & Parameter decoupling\\
\cite{liu2020secure,li2019fedmd} &  & \checkmark  &  & \checkmark & Transfer learning\\
\cite{li2020federated,shoham2019overcoming,yao2020continual} &  & \checkmark  &  & \checkmark & Regularization\\
\cite{NEURIPS2020_24389bfe,khodak2019adaptive,jiang2019improving,singhal2021federated} &  & \checkmark  &  & \checkmark & Meta-learning \\
\bottomrule
\end{tabular}%
\end{table}
\subsubsection{Discussion}
In addition to algorithmic fairness, client fairness is another concern in the FL community. Table \ref{tab: FAFL} enumerated various works on these two topics. Regarding client fairness, the single model approach focuses on smoothing data heterogeneity, where it is easy to implement and can be added to the general FL paradigm since it only needs modest modification. On the downside, the single model approach is less effective than personalized approaches in terms of capturing local data distribution and may be insufficient when the data distributions vary significantly between clients. Additionally, the single-model approach does not allow clients to customize their models.

\subsection{Discussion of fairness in FL}
There are two definitions of fairness in FL, \textit{client fairness} and \textit{algorithmic fairness}. \textit{Algorithmic fairness} has been extensively studied in centralized machine learning. These algorithms presuppose centralized access to data, however, one virtue of FL is data never leaves the device. This means neither the server nor any client gains centralized access to the training data. Therefore, generalizing the fair learning algorithms to FL is not trivial. On the one hand, data is stored locally in FL. The server cannot directly access the local data of clients. Hence, server-side debiasing is not a viable solution. On the other hand, debiasing on the client side is ineffective due to the inadequate data, which can hardly represent the global data distribution \cite{mcmahan2017communication}. There is no guarantee that model debiased with local data will generalize to the global distribution. The non-i.i.d data distributions further complicated this problem \cite{kairouz2019advances}.

\textit{Client fairness} is tailored to FL and stems from the non-i.i.d data. Each client sampled the training data from a distinct distribution. In this case, the vanilla FL protocol, \textit{FedAvg}, fails to train a model to fits clients' data distribution. Various methods have been proposed to alleviate this. From the data aspect, \cite{zhao2018federated, jeong2018communicationefficient,Hao_2021_CVPR} proposed to augment client data to yield an i.i.d dataset. \cite{9155494,yang2020federated,9141436} proposed to select participant clients to form a more homogeneous distribution. However, their methods did not consider the possible algorithmic fairness issues and may introduce bias to the model by choosing specific clients at a higher probability than others. From the model perspective, training different models for different clients seems a natural solution to the non-i.i.d. challenge. The core idea is to train global data collaboratively and then personalize it to local data distribution.

\section{Interactions between Privacy and Fairness}
\label{sec: interaction}
\begin{figure}[ht]
	\centering
	\includegraphics[width=8cm]{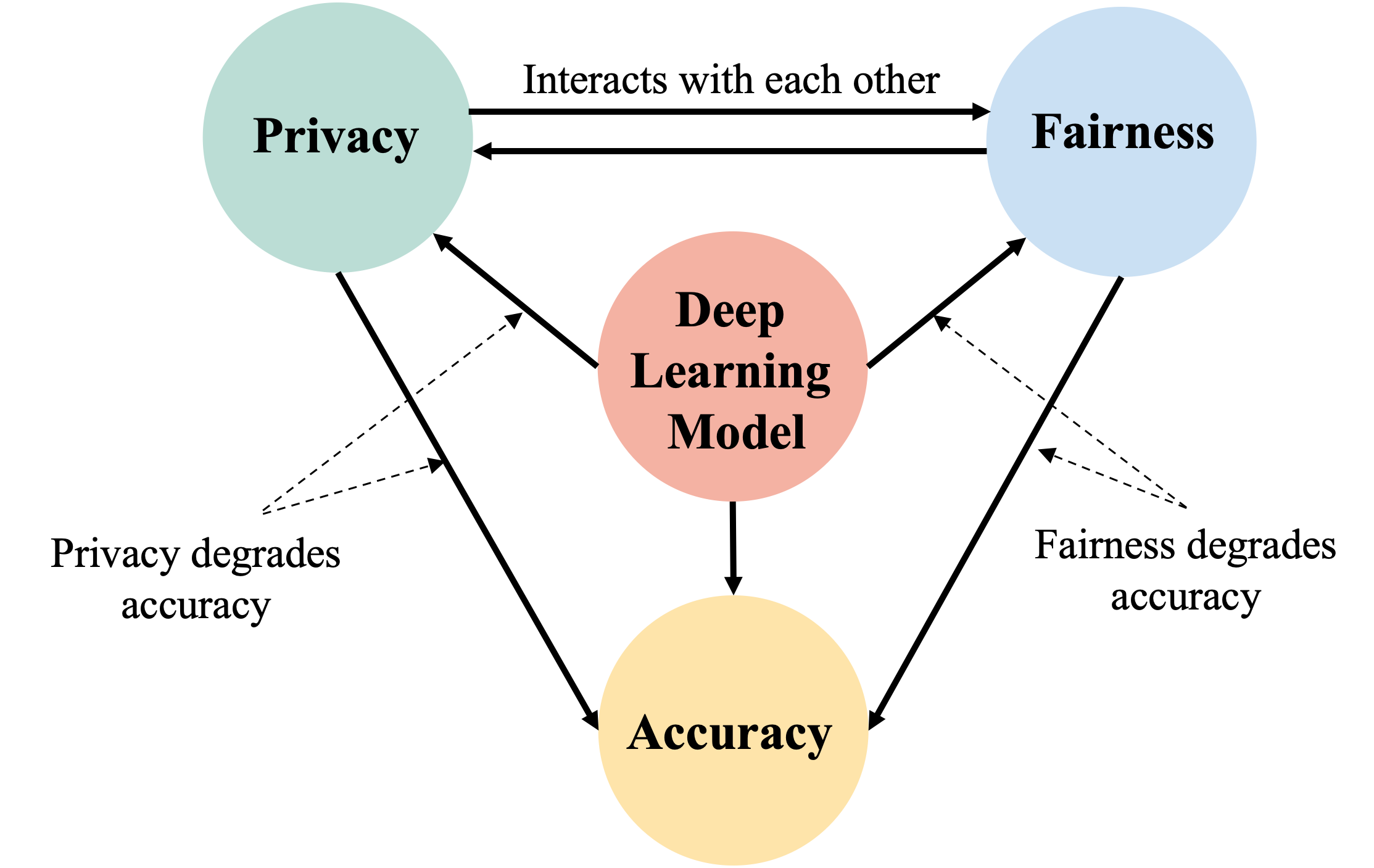}
	\caption{Privacy, fairness, and accuracy trade-offs in deep learning: 1) privacy comes at the cost of accuracy; 2) fairness comes at the cost of accuracy; and 3) privacy interacts with fairness \cite{zhang2021balancing}}
	\label{fig:trade-off}
\end{figure}

As shown in Fig \ref{fig:trade-off}, privacy and fairness are intertwined. On the one hand, \textit{fairness comes at the cost of privacy}. A fair model is trained to perform equally on data from different groups, which incurs overfit problems and consequently increases the privacy risk \cite{chang2021privacy}. On the other hand, \textit{privacy also harms fairness}. Several works \cite{bagdasaryan2019differential, Kuppam2020FairDM, tran2021decision, sanyal2022unfair} inconsistent reductions in accuracy caused by private mechanisms on classification \cite{farrand2020neither} and generative tasks \cite{ganev2022robin}. Due to the tension between fairness and privacy, researchers often need to make trade-offs between these two notions. The trade-off could be increasing privacy protection at the expense of fairness, i.e., adopting relaxed fairness notions instead of exact notions or the opposite way. Table \ref{tab:FP trade-off} gives various trade-offs between privacy and fairness. On the basis of the adopted privacy/fairness notions, the trade-offs can be divided into two categories. The first type sacrifices fairness for solid privacy protection, i.e. $epsilon$-DP. The second form prioritizes fairness over privacy and employs relaxed DP in order to accommodate exact fairness.
\begin{table}[]
\centering
\caption{Private and Fair Learning}
\label{tab:FP trade-off} \small
\setlength{\tabcolsep}{2pt} 
\renewcommand{\arraystretch}{1.4}
\begin{tabular}{cccccc}
\hline
\multirow{2}{*}{\textbf{Reference}} & \multirow{2}{*}{
\begin{tabular}[c]{@{}c@{}} \textbf{Privacy} \\ \textbf{notion}\end{tabular}} &  \multirow{2}{*}{
\begin{tabular}[c]{@{}c@{}} \textbf{Fairness} \\ \textbf{notion}\end{tabular}} & \multicolumn{2}{c}{\textbf{Techniques to achieve}} & \multirow{2}{*}{\begin{tabular}[c]{@{}c@{}} \textbf{Trade-off} \\ \textbf{type}\end{tabular}}\\ \cline{4-5}
& & & \textbf{Privacy } &  \textbf{Fairness} & \\ \hline

\cite{cummings2019compatibility} & $\epsilon$-DP & $\alpha$-Discrimination & \begin{tabular}[c]{@{}c@{}} Exponential\\ mechanism \end{tabular} &\begin{tabular}[c]{@{}c@{}} Minimize \\discrimination scores\end{tabular} & I\\ \hline
\cite{xu2019achieving} & $\epsilon$-DP &  \begin{tabular}[c]{@{}c@{}} Decision boundary \\ fairness\end{tabular} & \begin{tabular}[c]{@{}c@{}} Functional\\ mechanism \end{tabular} &  Fairness constraints & I \\ \hline

\cite{mozannar2020fair} & $\epsilon$-DP &  $\alpha$-Equal opportunity & Local DP & \begin{tabular}[c]{@{}c@{}}Post-processing\end{tabular}& I \\ \hline

\cite{lamy2019noise}& $\epsilon$-DP &  \begin{tabular}[c]{@{}c@{}}Equal odds \& \\ Demographic parity\end{tabular} & \begin{tabular}[c]{@{}c@{}}Class conditional\\ noise\end{tabular} & Fairness constraints & I \\ \hline

\cite{ding2020differentially} &  \begin{tabular}[c]{@{}c@{}}$\epsilon$-DP \&\\ $(\epsilon, \delta)$-DP\end{tabular} &  \begin{tabular}[c]{@{}c@{}}Decision boundary \\fairness \end{tabular} &  \begin{tabular}[c]{@{}c@{}}Functional \\mechanism \end{tabular}& Fairness constraints & I \\ \hline

\cite{jagielski2019differentially} & $(\epsilon, \delta)$-DP &  $\alpha$-Equal opportunity & \begin{tabular}[c]{@{}c@{}}Exponential\\ mechanism \\ \& Laplace noise\end{tabular}  & Fairness constraints & / \\ \hline

\cite{lowy2022stochastic} &$(\epsilon, \delta)$-DP & \begin{tabular}[c]{@{}c@{}}Equal odds \& \\ Demographic parity\end{tabular} & DP-SGDA & ERMI regularizer & II \\ \hline
\cite{esipova2022disparate} & $(\epsilon, \delta)$-DP & Excessive risk gap & DPSGD-Global-Adapt & Gradient correction & II \\ \hline

\cite{tran2021differentially} & \begin{tabular}[c]{@{}c@{}}  $(\alpha,\epsilon_p)$-\\Rényi DP\end{tabular} & \begin{tabular}[c]{@{}c@{}}Equal odds, \\ Accuracy parity\\ \& Demographic parity\end{tabular} & DP-SGD & Fairness constraints & II  \\ \hline
\cite{kilbertus2018blind} & / & Equal accuracy & MPC & Fairness constraints & II \\ \hline
\cite{gupta2018proxy} & / & Equal opportunity & Proxy attribute & Post-processing  & II \\ \hline
\cite{wang2020robust} & / & Demographic parity & Noisy attribute & Fairness constraints  & II \\ \hline
\cite{awasthi2020equalized} & / & Equal odds & Noisy attribute & Post-processing  & II \\ \hline
\multicolumn{6}{@{}l}{\small I: Trade fairness for privacy. Relaxing fairness notions to achieve purely DP} \\
\multicolumn{6}{@{}l}{\small II: Trade privacy for fairness. Adopting relaxed DP notion to accommodate exact fairness}
\end{tabular}
\end{table}

\subsection{Privacy Degrades Fairness}
Currently, there are two approaches to privately train a fair model: the cryptography approach and the DP approach. In the first line of works, Veale et al. \cite{veale2017fairer} suggested storing protected characteristics in a trusted third party to protect privacy. The third-party will perform discrimination discovery and incorporate fairness constraints into model-building. Kilbertus et al. \cite{kilbertus2018blind} relaxed the assumption of a trusted third party and proposed to train a fair model with encrypted sensitive attributes \cite{mozannar2020fair}. The cryptography approach is guaranteed to provide comparable performance as the non-private alternative at the cost of communication overhead. The trade-offs between privacy and efficiency are the main concern in this category.

However, as argued by \cite{jagielski2019differentially}, the cryptographic approach does not guarantee privacy at the inference stage. It only ensures the training data remain private during training and cannot prevent the adversary from inferring training samples from the neural network parameters \cite{zhao2020idlg,song2017machine}. On the contrary, DP guarantees a fair model will not leak anything beyond what could be carried out from "population level" correlations. As such, the majority of works focus on learning fair, and DP model \cite{jagielski2019differentially,cummings2019compatibility,bagdasaryan2019differential,xu2019achieving}. Thus, this subsection focus on DP as the privacy-preserving techniques. 
\subsubsection{Empirical Findings}
The impact of privacy on fairness was initially observed in empirical studies. Bagdasaryan et al. \cite{bagdasaryan2019differential} first observed that the reduction in accuracy caused by deep DP models negatively impacts underrepresented subgroups disproportionately. DP-SGD strengthens the model's "bias" toward the most prominent features of the distribution that is being learned. Kuppam et al. \cite{Kuppam2020FairDM} reached a similar conclusion when they examined the effects of DP on fairness in three real-world tasks involving sensitive public data. When the noise added by a private algorithm is negligible in relation to the underlying statistics, the costs of adopting a private technique may be minor. When stronger privacy is implemented or when a task entails a small population, significant disparities may emerge. Farrand et al. \cite{farrand2020neither} demonstrated that even minor differences and weak privacy protections could result in disparate outcomes. Ganev et al. \cite{ganev2022robin} shifted the emphasis to generative models and tabular synthetic data. Three DP generative models PrivBayes \cite{zhang2017privbayes}, DP-WGAN \cite{alzantotdifferential}, and PATE-GAN \cite{jordon2019pate}, were involved. They witnessed a disparate effect on the accuracy of classifiers trained on synthetic data generated by all generative models. The losses are greater and/or more dispersed for underrepresented groups. Uniyal et al. \cite{uniyal2021dp} compared DP-SGD and Private Aggregation of Teacher Ensembles (PATE) \cite{papernot2016semi}, an alternative DP mechanism for discreetly training a deep neural network, in terms of fairness. They discovered that PATE has a disparate effect, but it is considerably less severe than DP-SGD.
\subsubsection{Theoretical Explanations}
Several works have attempted to determine the mechanism underlying the well-known relationship between privacy and unfairness. Bagdasaryan et al. \cite{bagdasaryan2019differential} associated the impact with the gradient clipping operation in DP-SGD. During training, the model generates larger gradients for samples from underrepresented subgroups; consequently, clipping slows their learning rate. Therefore, the model learns less from the underrepresented subgroups, and its performance on those subgroups is negatively impacted more. Tran et al. \cite{tran2021differentially} conducted an in-depth study into this phenomenon with output perturbation \cite{chaudhuri2011differentially} and DP-SGD as the private mechanism. By measuring fairness with \textit{excessive risk gap}, Tran et al. proved that output perturbation mechanisms incur unfairness when the local curvatures of the loss functions of different groups differ substantially. For DP-SGD, Tran et al. found that the clipping bound, the norm of inputs, and the group’s distance to the decision boundary collectively contributed to the unfairness raised by DP-SGD. Esipova et al. \cite{esipova2022disparate} examined the same issue from the gradient perspective. They proved that the gradient misalignment caused by DP-SGD is the main reason for unfairness. If the clipping operation disproportionately and sufficiently increases the direction error for group $a$ relative to group $b$, then group $a$ incurs larger excessive risk due to gradient misalignment.

\subsubsection{Mitigation Strategies}
Diverse methods have been proposed to mitigate the effect of private mechanisms on fairness. Xu et al. \cite{10.1145/3447548.3467268} proposed DP-SGD-F, a variant of DP-SGD that reduces the divergent impact on different populations. By adaptively designating clipping bounds for each group, DP-SGD-F achieves a level of privacy proportional to each group's utility-privacy trade-off. For the group whose clipping bias is greater (due to large gradients), a larger clipping bound is adopted to mitigate for their greater privacy cost. Tran et al. \cite{tran2021differentially} formulated a regularized optimization problem that minimizes empirical loss while satisfying two additional constraints. The first constraint equalizes the averaged non-private and private gradients, while the second constraint penalizes the difference between the local curvatures of distinct groups' loss functions. Esipova et al. \cite{esipova2022disparate} modified DP-SGD and developed DP-SGD-Global-Adapt to preserve gradient direction. It is assumed that a hyperparameter $Z$ is the upper bound for most gradients. Gradients less than $Z$ are uniformly scaled, whereas gradients greater than $Z$ are trimmed to $Z$.

\subsection{Fairness Increases Privacy Risk}
Fairness in turn presents challenges for privacy mechanisms. Chang and Shokri \cite{chang2021privacy} observed an elevated privacy risk for underprivileged subgroups in a fair model. In order to ensure fairness, the model must perform equally well for all subgroups. However, limited data availability for underprivileged subgroups can lead to overfitting of the training data for unprivileged subgroups in a fair model, thereby increasing the privacy risk. Previous works on fairness-aware machine learning often assume that the sensitive features are reliable and accessible. This assumption unavoidably introduces privacy risks. However, achieving precise notions of fairness, such as demographic parity, becomes unattainable without access to sensitive attributes, specifically the membership information of sensitive groups. To address this issue, several techniques have been proposed to safeguard the privacy of sensitive attributes during the training of fair models.

\subsubsection{Training Fair Models with Noisy Representation}
Researchers in this field train approximately fair models using noisy sensitive attributes to protect privacy. Gupta et al. \cite{gupta2018proxy} substituted protected groups with proxy groups. To achieve fairness, the proxy groups need to align with the true positive group and even overlap with the ground-truth groups. Thus, the fairness guarantee comes at the cost of privacy. Several studies have explored fairness with imperfect group information. \cite{kallus2020assessing,wang2020robust,lamy2019noise, awasthi2020equalized}. Lamy et al. \cite{lamy2019noise} introduced a mutual contaminated model to simulate a noisy distribution with corrupted attributes. Under this framework, they demonstrated that the fairness constraint on the clean distribution is equivalent to a scaled fairness constraint on the noisy distribution. To protect the privacy of sensitive attributes, they added class conditional noise to release the noisy dataset. Awasthi et al. \cite{awasthi2020equalized} addressed the challenging problem of training a fair model with perturbed sensitive attribute values, where each attribute is independently flipped to its complementary value with probability $\gamma$. They identified conditions on the perturbation under which the classifier, denoted as $\hat{Y}$, obtained by Hardt et al.'s method \cite{hardt2016equality}, is fairer than the vanilla classifier, denoted as $\tilde{Y}$, trained on accurate attributes. They further provided a formal guarantee of effectiveness under the necessary conditions. Wang et al. \cite{wang2020robust} trained a fair binary classifier based on a noisy label $\hat G \in \{1,...,\hat{m}\}$, i.e., $\hat G $ could be \textit{"country of residence"} as a noisy representation of the true group labels $G =$ \textit{"language spoken at home"}. 

\subsubsection{Training Fair Models with DP}
Works in this area protect privacy by adding noise to the private characteristic \cite{tran2021differentially}. The trade-off between privacy and fairness depends on the amount of noise added, with no noise and excessive noise representing the two extremes. In the case of no noise, the model's performance remains unaffected but could lead to information breaches. Conversely, high levels of noise are effective in preserving privacy but can compromise the model's utility. Tran et al. \cite{tran2021differentially} proposed a constrained optimization problem to address both private and fair learning tasks. Their framework ensures $(\alpha,\epsilon_p)$-Rényi DP \cite{mironov2017renyi} for the sensitive attributes by solving the constrained problem with DP-SGD. Jagielski et al. \cite{jagielski2019differentially} extended Agarwal et al.'s approach \cite{agarwal2018reductions} by incorporating privacy considerations. They formulated a two-player zero-sum game, played between a "learner" and an "auditor," to derive a fair classifier. Laplacian noise \cite{10.1007/11681878_14} and the exponential mechanism \cite{mcsherry2007mechanism} were utilized separately for the "learner" and the "auditor". As a result, the learned model satisfies $(\epsilon,\delta)$-DP and achieves equalized odds.

\subsection{Fair and Private FL}
In centralized machine learning, one entails centralized access to training data (either the true data or noisy data). However, this is invalid in FL, where neither the server nor clients have access to others' data. Therefore, one cannot simply apply centralized fair learning algorithms in FL tasks. This raises a question: \textit{How can we promote algorithmic fairness in FL without accessing clients' data in FL}? Several studies made progress in response to this challenge.

\textbf{Using a surrogate model to preserve privacy}. Padala et al. \cite{padala2021federated} tried to satisfy both $(\epsilon,\delta)$-local DP and demographic fairness through a fair and private FL framework. To circumvent the access restriction, they decomposed the learning into two phases. First, each client learns a fair and accurate model on a local dataset, where the fairness constraint acts as a regularization term in the loss function. Then, every client trains a surrogate model to match the fair predictions from the first model with a DP guarantee. Finally, only the surrogate model is communicated to the server. 

\textbf{Privacy through secure aggregation}. Zhang et al. \cite{zhang2020fairfl} investigated classification problems in FL through multiple goal optimization problems with privacy constraints. The objective is to minimize the accuracy loss and the discrimination risk. To this end, a team Markov game was designed to select participating clients at each communication round. In each round, clients decide whether or not to participate based on the global model’s state, which is characterized by bias level and accuracy. Further, a secure aggregation protocol is designed to estimate the global model's status based on polynomial interpolation \cite{karnin1983secret} for privacy concerns. Under this protocol, the server is able to calculate the discrimination status without accessing the local data. 

\textbf{Achieve fairness based on statistics}. Gálvez et al. \cite{Galvez2021EnforcingFI} formulated a constrained optimization problem that is solved by the differential multiplier. Local statistics are provided to the server for debiasing the global model. To further protect privacy, client updates are clipped and perturbed by Gaussian noise before being sent to the server. Finally, their solution is able to provide the approximate group fairness notion over multiple attributes and $(\epsilon,\delta)$-DP.

\textbf{Fairness through agnostic learning}. Shifts in distribution is one source of bias in FL. The global model is trained on the data of all clients (source distribution), but each client’s local data distribution (target distribution) may differ. When deployed to the client, unfavorable outcomes occur. Du et al. \cite{du2021fairness} proposed treating the client data distribution in an agnostic way. An adversary generates any possible unknown local data distribution to maximize the loss, while the learner aims to optimize the accuracy and fairness. 

\textbf{Calculate fairness violations locally}. Chu et al. \cite{Chu2021FedFairTF} formulated a constraint optimization problem to learn a fair and private model in FL. Each client locally calculates fairness violations to avoid impinging on the data privacy of any client. Chu et al. \cite{Chu2021FedFairTF} further optimized this method by aggregating fairness constraints to better estimate the true fairness violation for all data. 

Although some fair FL algorithms do not directly access the training data \cite{du2021fairness,Chu2021FedFairTF}, faithfully sharing the model/gradients in FL could incur privacy leakage risks. The privacy breach could happen during the training or inference stage. The attack could be carried out by either the server or the clients \cite{yin2021comprehensive}. For example, an honest-but-curious server can lunch a reconstruction attack \cite{zhu2019deep} to recover the private data from the gradients uploaded by the victim client. However, the main challenge to training a fair model in FL is restricted data access, e.g., data never leaving local devices, which is an under-investigated topic in FL literature. In the case of the adversary clients/server in FL, some privacy-preserving techniques, such as DP, can be combined with the aforementioned fair FL approaches to prevent privacy leakage.

\subsection{Discussion of Privacy and Fairness Interactions}
The complex interactions between privacy and fairness have been thoroughly examined and documented in various studies. These investigations highlight the intricate trade-offs and challenges that arise when attempting to simultaneously address both privacy and fairness objectives \cite{cummings2019compatibility}.

The impact of privacy and fairness on each other is indeed bilateral. In one scenario, privacy measures can degrade fairness. For instance, in widely-used privacy mechanisms like DP-SGD, to protect privacy, the algorithm clips and adds noise to the gradients. However, due to the scarcity of data for certain groups, these modifications can disproportionately affect underrepresented groups, exacerbating unfairness. Therefore, the implementation of DP can inadvertently worsen existing unfairness by disproportionately impacting certain groups. 

In another case, fairness can increase privacy risks. To achieve fairness, it may be necessary to collect additional demographic information about users, even if it is irrelevant to the task at hand. This data collection is aimed at guiding modifications to the model, such as addressing inconsistent responses or removing discrimination in statistical models \cite{voigt2017eu,goodman2016step,liobait2016UsingSP}. However, the collection of such sensitive information raises privacy concerns, as it expands the scope of data being collected and potentially increases the risk of privacy breaches.

In the context of Federated Learning (FL), the cooperative game between clients and the server adds complexity to the privacy and fairness challenges. FL introduces new privacy attack surfaces, as discussed in Section 3, where potential malicious participants can actively or passively infer the private data of other clients. Consequently, securing private information in FL requires even stronger privacy protection measures compared to the centralized setting. Merely protecting group membership is insufficient to address the privacy risks in FL. Furthermore, the non-i.i.d. (non-independent and identically distributed) nature of FL poses another challenge. In a typical FL system, clients' data are sampled from different distributions, leading to data heterogeneity. A model that achieves fairness within the local distribution of each client is not guaranteed to perform unbiasedly on a global scale. The non-i.i.d. issue also introduces potential fairness concerns at the client level, as the performance of the model can vary significantly among clients. It is crucial to address this variation and ensure fairness across all participating clients in FL. The challenge lies in training a fair model in FL without violating the data access restrictions imposed by each client. Finding methods to mitigate the fairness issues arising from the non-i.i.d. nature of the data while respecting the privacy and data access constraints in FL remains a challenging task.

\section{Open Research Directions}
\label{sec: open direction}
The research community has made fruitful progress in privacy and fairness in FL. However, throughout this survey, we found this field still faces several challenges that need to be solved.
\begin{itemize}
    \item \textbf{Trade-offs between Privacy and Fairness}. The interaction between privacy and fairness is an under-studied topic. Existing works have focused on exploring the two notions in isolation, either focused on privacy-preserving machine learning \cite{xu2021privacy} or on paradigms that respect fairness \cite{shi2021survey}. However, as demonstrated by several studies \cite{chang2021privacy,bagdasaryan2019differential,Kuppam2020FairDM}, privacy and fairness may compete with each other.  In the realm of FL, challenges and opportunities coexist. On the one hand, restricted information and non-i.i.d distribution complicate the problem settings. On the other hand, the flexibility of the FL paradigm may enable more possible solutions. For instance, the personalized model \cite{kulkarni2020survey,tan2021towards,zhu2021federated} has been widely used in FL to address statistical challenges by assigning clients personalized models. We may combine privacy and personalized models to achieve a better trade-off between privacy, fairness, and utility. Thus, we believe it is worth examining the trade-offs between privacy and fairness in FL.   \item \textbf{The Compatibility of Fairness and DP}. We believe it would be worth investigating techniques that simultaneously accommodate fairness and DP. As pointed out in Dwork et al.'s \cite{dwork2012fairness} work, given a carefully designed distance metric, it is possible to achieve individual fairness through $\epsilon$-DP. Two characteristics of FL make individual fairness a superior choice over group fairness: 1) Data distribution in FL may vary significantly between clients, and individual fairness is more suitable in such cases. Since it is defined at the sample level, thus, it generates better than group notions when addressing new samples which may be distinct from those in the training set; 2) The restricted access to information in FL lends itself more to individual fairness because individual fairness relies on a Lipschitz continual prediction model and does not require access to demographic data. This perfectly fits the FL setting. 
    
    \item \textbf{How can one satisfy fairness at both the algorithm and client levels in FL?} The majority of studies on fairness in FL focus on promoting fairness at the client level. However, client-level fairness does not necessarily imply algorithmic fairness. Consider a scenario where multiple companies (clients) collaborate to train a credit card approval model. Consumer demographic compositions vary between each company. Although a federated model trained subject to client-level fairness constraints might handle the different companies fairly, the model could still be biased towards sensitive attributes (such as race or educational background). This raises a question: \textit{How can one satisfy fairness at both the algorithm and the client levels while preserving privacy in FL?}. 
\end{itemize}
\section{Conclusion}
\label{sec: conclusion} 
In this article, we conducted a detailed survey of data privacy and model fairness issues in FL. Uniquely, we also documented the interactions between privacy and fairness from the perspective of trade-offs. In terms of privacy in FL, we first reviewed privacy attacks in FL. Then, we presented three kinds of privacy-preserving techniques. Regarding fairness, we first analyzed the possible sources of bias and how bias can be introduced on both the client and server sides. Following a review of the notions of fairness adopted in machine learning and those originating from FL, a discussion of the various fairness-aware FL algorithms is presented. The last part of the survey focused on the interactions between privacy and fairness. We identified three relations in the general context and further listed possible solutions to achieve both fair and private FL.
\begin{acks}
   This paper is supported by the Australian Research Council Discovery DP200100946 and DP230100246, and NSF under grants III-1763325, III-1909323,III-2106758, and SaTC-1930941. 
\end{acks}
\bibliographystyle{ACM-Reference-Format}
\bibliography{reference}
\end{document}